\documentclass[11pt,letterpaper]{article}

\usepackage[utf8]{inputenc}
\usepackage[T1]{fontenc}
\usepackage{lmodern}
\usepackage{microtype}
\usepackage[margin=1in]{geometry}
\usepackage{graphicx}
\usepackage{booktabs}
\usepackage{array}
\usepackage{longtable}
\usepackage{enumitem}
\usepackage{xcolor}
\usepackage{amsmath,amssymb}
\usepackage{hyperref}
\usepackage{cleveref}  
\usepackage{tikz}
\usetikzlibrary{shapes,arrows,arrows.meta,positioning,fit,backgrounds,shadows}
\usepackage{pgfplots}
\pgfplotsset{compat=1.17}
\usepgfplotslibrary{groupplots}
\usepackage{orcidlink}
\usepackage{tcolorbox}
\tcbuselibrary{skins}

\definecolor{pblue}{HTML}{275D8C}   
\definecolor{pteal}{HTML}{137A63}   
\definecolor{pgreen}{HTML}{2E7D32}  
\definecolor{pamber}{HTML}{B45309}  
\definecolor{pred}{HTML}{B3382C}    
\definecolor{ppurple}{HTML}{6B4C9A} 
\definecolor{pslate}{HTML}{4A5568}  
\usepackage{listings}
\usepackage[numbers,sort&compress]{natbib}
\tcbset{
  paperbox/.style 2 args={
    enhanced,
    colback=#1!4!white,
    colframe=#1!75!black,
    colbacktitle=#1!10!white,
    coltitle=#1!50!black,
    fonttitle=\bfseries\sffamily\small,
    title=#2,
    arc=1.2mm,
    boxrule=0.5pt,
    leftrule=2.5pt,
    titlerule=0pt,
    toptitle=0.8mm, bottomtitle=0.8mm,
    left=2.5mm, right=2.5mm, top=1.5mm, bottom=1.5mm,
    before skip=8pt, after skip=8pt,
  }
}

\newtcolorbox{keytakeaways}{
  paperbox={pblue}{Key Takeaways}
}

\newtcolorbox{definition}[1][]{
  paperbox={pslate}{#1}
}

\newtcolorbox{warningbox}{
  paperbox={pamber}{Warning}
}

\newtcolorbox{promptbox}[1][Prompt]{
  paperbox={pteal}{#1},
  fonttitle=\bfseries\sffamily\small\ttfamily
}

\newtcolorbox{responsebox}[1][Response]{
  paperbox={pblue}{#1},
  fonttitle=\bfseries\sffamily\small\ttfamily
}

\newtcolorbox{badexample}[1][Failure Case]{
  paperbox={pred}{{\textcolor{pred}{$\times$}} #1}
}

\newtcolorbox{goodexample}[1][Corrected Case]{
  paperbox={pgreen}{{\textcolor{pgreen}{$\checkmark$}} #1}
}

\newtcolorbox{empirical}[1][Empirical Finding]{
  paperbox={ppurple}{#1},
  fonttitle=\bfseries\sffamily\small\itshape
}

\hypersetup{
  colorlinks=true,
  linkcolor=blue!70!black,
  citecolor=green!50!black,
  urlcolor=blue!70!black,
  pdftitle={When Generic Prompt Improvements Hurt: Evaluation-Driven Iteration for LLM Applications},
  pdfauthor={Daniel Commey},
  pdfsubject={LLM Evaluation},
  pdfkeywords={large language models, evaluation, benchmarks, metrics, RAG, LLM-as-judge}
}

\title{%
  \textbf{When Generic Prompt Improvements Hurt:}\\[0.5em]
  \Large Evaluation-Driven Iteration for LLM Applications
}

\author{%
  Daniel Commey\,\orcidlink{0000-0001-5759-918X}
}

\date{}

\begin{document}

\maketitle


\begin{abstract}
Evaluating Large Language Model (LLM) applications differs from conventional software testing because outputs are probabilistic, semantically variable, and sensitive to prompt and model changes. This technical report proposes the Minimum Viable Evaluation Suite (MVES), an audit-oriented structure for application-level LLM evaluation. MVES links application categories to failure modes, metrics, required artifacts, and validation evidence across general LLM applications, retrieval-augmented systems, and agentic workflows.

We pair the framework with a reproducible local evaluation harness covering structured extraction, RAG citation/content-compliance, and instruction-following checks. Using Ollama with Llama 3 8B Instruct and Qwen 2.5 7B Instruct, we evaluate five prompt conditions over expanded 30-case-per-suite ablations. The results show that, in the tested local conditions, generic prompt additions do not produce monotonic improvements: stronger output-contract prompts improve strict extraction for both models, while RAG citation/content-compliance declines under some generic-rule conditions. The largest observed decline occurs for Qwen 2.5 on RAG when generic rules are appended to the user prompt, from 26/30 to 9/30.

These findings support evaluation-driven prompt iteration: prompt changes should be treated as potential regression risks and tested against task-specific suites before deployment. The accompanying repository contains the test suites, prompt variants, evaluation harness, raw result logs, and scripts needed to reproduce the reported local ablations.
\end{abstract}


\paragraph{Contributions.}
This report makes four contributions. First, it proposes \textbf{MVES}, an audit-oriented structure that links LLM application evaluation to failure modes, required artifacts, metrics, and validation evidence. Second, it provides a \textbf{reproducible local evaluation harness} for structured extraction, RAG citation/content-compliance, and instruction-following checks. Third, it reports a \textbf{prompt-ablation study} showing that generic prompt additions can improve one task contract while degrading another, with effects varying by model and prompt placement. Fourth, it consolidates \textbf{practical evaluation guidance} for test-set design, metric selection, LLM-as-judge calibration, and production monitoring in a form intended to support reproducible application-level evaluation.

\newpage
\tableofcontents
\newpage


\section{Introduction}
\label{sec:introduction}

Large Language Models (LLMs) are now widely deployed in production applications. Organizations use LLMs for customer support, document summarization, code generation, knowledge retrieval, and other text and decision-support tasks. Yet evaluating these systems remains difficult in practice. Traditional software testing, which assumes deterministic outputs for given inputs, does not translate directly to LLM-powered applications.

\subsection{Why LLM Evaluation Differs from Traditional Testing}

Consider a conventional API: given a well-formed request, the response is deterministic and can be validated against an expected output. LLM applications violate nearly every assumption underlying this paradigm.

The first challenge is \textbf{non-determinism}. Even with identical prompts and temperature set to zero, LLMs may produce slightly different outputs across inference calls due to hardware numerics, sampling implementations, and batching effects~\citep{ouyang2022training}. This variability means that exact-match testing, the foundation of traditional software verification, becomes unreliable.

The second challenge is \textbf{output space complexity}. Natural language responses can be semantically equivalent while being lexically distinct. The statements ``The capital of France is Paris'' and ``Paris serves as France's capital'' convey identical information but differ textually. Evaluating semantic equivalence requires more sophisticated methods than string comparison.

The third challenge involves \textbf{implicit specifications}. User expectations for ``good'' responses are often context-dependent and difficult to formalize. A concise answer may be preferred in one context and insufficient in another. Unlike APIs with explicit schemas, judgments of LLM output quality often depend on user goals, domain conventions, and risk tolerance.

Finally, \textbf{model churn} complicates evaluation over time. LLM providers frequently update their models, sometimes without explicit versioning. A system that worked yesterday may behave differently today~\citep{sainz2023nlp}. Continuous evaluation becomes necessary to detect regressions introduced by upstream model changes.

\subsection{Key Risks in LLM Applications}

Insufficient evaluation exposes applications to several categories of risk that must be addressed before deployment.

\textbf{Hallucination} refers to the generation of plausible-sounding but factually incorrect statements. This phenomenon has been extensively documented in the literature~\citep{ji2023hallucination, lin2022truthfulqa}. In high-stakes domains such as healthcare or legal advice, hallucinations can cause material harm. Evaluation must specifically test for factual accuracy against verified sources.

\textbf{Safety violations} occur when LLMs produce harmful, biased, or inappropriate content. Without appropriate guardrails, models may respond to adversarial prompts in dangerous ways. Red-teaming and adversarial testing are essential to surface these failure modes before deployment~\citep{perez2022red, ganguli2022red}. Safety evaluation should cover toxic content, privacy violations, and refusal of genuinely harmful requests.

\textbf{Prompt drift} emerges as prompts are iteratively refined during development. Subtle changes to system prompts or few-shot examples may have unintended effects on unrelated behaviors. Without regression testing, such regressions go undetected until reported by users. Comprehensive test suites help ensure that improvements in one area do not cause degradation in others.

\textbf{Distribution shift} occurs when production inputs differ from development test cases. Users may phrase requests in unexpected ways, submit adversarial inputs, or use the system for unanticipated purposes. Evaluation should include realistic samples from production, not just curated examples that developers find convenient.

\subsection{What This Report Provides}

This report contributes two linked artifacts. First, it proposes MVES, an audit-oriented framework that connects evaluation components to specific failure modes, required artifacts, and validation evidence. Second, it provides a local evaluation harness with hand-authored seed suites and deterministic expansion to 30 cases per task in the reported ablation. The harness is small by design: it illustrates evaluation-driven iteration and artifact structure rather than estimating population-level model reliability.

The local ablation illustrates a failure pattern: generic prompt improvements are not necessarily monotonic across task contracts. In expanded 30-case-per-suite runs, stronger output-contract prompts improved strict extraction, but RAG citation/content-compliance declined under several generic-rule conditions. A five-condition ablation suggests that the effect is not reducible to a short system wrapper; generic rules and non-conflicting guidance can affect structured-output, RAG, and instruction-following checks differently across models.

The result suggests that prompt changes should be validated against task-specific test suites rather than assumed beneficial. The evaluation loop described in Section~\ref{sec:eval-loop} operationalizes this insight.

\paragraph{Scope.} This report is not intended to rank LLMs, establish universal prompt-engineering rules, or provide a comprehensive survey of LLM evaluation tools. The empirical study is a controlled local demonstration of prompt-regression behavior: its purpose is to show how small prompt changes can affect different task contracts differently, and how an auditable evaluation suite can expose those changes before deployment.

\paragraph{Artifacts.} The accompanying repository contains the evaluation harness, test suites, prompt variants, raw result logs, and scripts used to reproduce the reported ablations: \url{https://github.com/dcommey/llm-eval-benchmarking}


\section{The Evaluation Loop}
\label{sec:eval-loop}

Effective LLM evaluation follows a structured iteration cycle. This section introduces a four-phase workflow that serves as the organizing principle for the remainder of this paper.

\subsection{The Core Workflow}

Traditional software testing verifies outputs against known correct answers. LLM applications complicate this model because outputs are often unstructured, subjective, or context-dependent. Nevertheless, a disciplined evaluation process remains essential.

The evaluation loop consists of four phases applied repeatedly throughout development. In the \textbf{Define} phase, teams articulate quality requirements in testable terms. What constitutes acceptable output for this application? What failures are most costly? In the \textbf{Test} phase, the system is evaluated against a curated suite of inputs with known properties. In the \textbf{Diagnose} phase, failures are categorized to identify systematic patterns. In the \textbf{Fix} phase, prompts, retrieval logic, or model selection are adjusted based on the diagnosis. The cycle then repeats.

This workflow differs from one-time benchmarking in two important ways. First, it treats evaluation as continuous rather than gated. Each prompt change or model update triggers re-evaluation. Second, it emphasizes failure analysis over aggregate metrics. Understanding why a case failed matters more than computing a single accuracy number.

\subsection{Translating Requirements into Tests}

Many LLM applications have implicit quality requirements that were never formalized. A customer support chatbot should be ``helpful'' and ``accurate,'' but what does this mean in testable terms?

The translation process involves decomposing high-level requirements into concrete properties. Consider a chatbot that answers questions using a knowledge base. ``Helpful'' might translate to: responds within 5 seconds, provides actionable next steps, and avoids jargon. ``Accurate'' might translate to: all factual claims are supported by retrieved documents, dates and numbers match source material, and the system declines to answer when sources are insufficient.

Each property then becomes a check in the evaluation harness. Some checks are fully automated (response latency, JSON validity, citation presence). Others require human judgment or LLM-as-judge scoring (helpfulness, clarity). The goal is to maximize coverage with automated checks while reserving human review for genuinely subjective dimensions.

\subsection{The Role of Golden Sets}

A golden set is a curated collection of inputs with known-good outputs or annotated properties. Unlike exhaustive test suites, golden sets prioritize coverage of failure modes over volume.

Effective golden sets share several characteristics. They include representative examples from each major use case. They contain adversarial inputs designed to trigger known failure modes. They are version-controlled alongside the prompt templates they evaluate. They are small enough to run on every change (50-200 cases) but large enough to detect regressions with statistical confidence.

The test suites used in Section~\ref{sec:experiments} demonstrate this approach. The repository contains hand-authored seed suites for extraction, RAG citation/content-compliance, and instruction following; the reported ablation expands these seeds deterministically to 30 cases per task. The suites are kept small so they can run on every change; they are not sized to estimate production reliability.

\subsection{Iteration in Practice}

The evaluation loop accelerates development by providing immediate feedback on prompt changes. Without it, teams often discover failures in production, leading to reactive fixes and degraded user trust.

Consider a scenario where a RAG application begins generating unsupported claims after a prompt update. With an evaluation loop in place, the team runs the golden set before deployment and observes a drop in citation compliance. They diagnose the issue: the new prompt's emphasis on helpfulness led the model to answer confidently even when sources were insufficient. They fix it by adding an explicit instruction to decline when evidence is lacking. They re-test to confirm the fix worked without introducing new regressions.

This scenario illustrates why evaluation-driven iteration is more reliable than intuition-based prompt engineering. The loop catches regressions that would otherwise reach users, and the diagnosis step provides actionable insight rather than vague failure signals.

\subsection{Connecting Offline and Online Evaluation}

Offline evaluation (golden sets, unit tests) and online evaluation (production monitoring, A/B tests) serve complementary roles. Offline evaluation catches known failure modes before deployment. Online evaluation detects novel failures and distribution shifts that offline suites did not anticipate.

The metrics defined in Section~\ref{sec:metrics-scoring} bridge these two contexts. The same checks that run in the evaluation harness (JSON validity, citation compliance, format constraints) can be logged in production. When online metrics diverge from offline baselines, the system signals potential regressions for investigation.

The remaining sections define the evaluation components and report the ablation study. Sections~\ref{sec:testset-design} through \ref{sec:metrics-scoring} address the Define phase. Sections~\ref{sec:eval-methods} and \ref{sec:llm-judges} address the Test phase. Section~\ref{sec:failure-modes} addresses the Diagnose phase. Section~\ref{sec:experiments} demonstrates the complete cycle with concrete before-and-after results.

\begin{keytakeaways}
The evaluation loop (Define-Test-Diagnose-Fix) replaces ad-hoc testing with systematic iteration. Golden sets catch regressions before deployment. Offline metrics validate changes rapidly, while online monitoring detects drift in production.
\end{keytakeaways}



\begin{figure}[htbp]
\centering
\resizebox{0.95\textwidth}{!}{%
\begin{tikzpicture}[
    node distance=1.0cm and 1.2cm,
    every node/.style={font=\small},
    stagebox/.style={
        rectangle,
        rounded corners=2.5pt,
        minimum height=0.95cm,
        minimum width=2.1cm,
        text centered,
        align=center,
        line width=0.7pt,
        drop shadow={shadow xshift=0.5mm, shadow yshift=-0.5mm,
                     fill=black, opacity=0.12},
    },
    databox/.style={
        stagebox,
        draw=pteal!80!black,
        fill=pteal!8,
        minimum height=0.8cm,
        minimum width=1.9cm,
    },
    box/.style={
        stagebox,
        draw=pblue!80!black,
        fill=pblue!8,
    },
    evalbox/.style={
        stagebox,
        draw=pamber!80!black,
        fill=pamber!8,
    },
    outputbox/.style={
        stagebox,
        draw=ppurple!80!black,
        fill=ppurple!8,
    },
    arrow/.style={
        -{Stealth[length=2.4mm, width=1.8mm]},
        line width=0.8pt,
        draw=pslate!75,
        rounded corners=2pt,
    },
    stagelabel/.style={
        font=\footnotesize\bfseries\sffamily,
    }
]

\node[databox] (testset) {Test Set};

\node[databox, below=0.6cm of testset] (context) {Retrieved\\Context};

\node[box, right=1.2cm of testset] (llm) {LLM\\Application};

\node[box, right=1.2cm of llm] (output) {Generated\\Output};

\node[evalbox, above right=0.4cm and 1.2cm of output] (automated) {Automated\\Metrics};
\node[evalbox, right=1.2cm of output] (llmjudge) {LLM-as-Judge};
\node[evalbox, below right=0.4cm and 1.2cm of output] (human) {Human Eval};

\node[outputbox, right=1.2cm of llmjudge] (scores) {Quality\\Scores};

\node[outputbox, right=1.0cm of scores] (dashboard) {Dashboard};

\begin{scope}[on background layer]
    \node[fit=(testset)(context), rounded corners=4pt,
          fill=pteal!4, draw=pteal!30, dashed, inner sep=2.5mm] (g-inputs) {};
    \node[fit=(automated)(llmjudge)(human), rounded corners=4pt,
          fill=pamber!4, draw=pamber!30, dashed, inner sep=2.5mm] (g-eval) {};
\end{scope}

\draw[arrow] (testset) -- (llm);
\draw[arrow] (context) -| (llm);
\draw[arrow] (llm) -- (output);
\draw[arrow] (output) -- (automated);
\draw[arrow] (output) -- (llmjudge);
\draw[arrow] (output) -- (human);
\draw[arrow] (automated) -- (scores);
\draw[arrow] (llmjudge) -- (scores);
\draw[arrow] (human) -- (scores);
\draw[arrow] (scores) -- (dashboard);

\draw[arrow, dashed, draw=pred!70, bend left=35]
    (dashboard.south) to node[below, font=\footnotesize\itshape, text=pred!80!black] {Iterate}
    (g-inputs.south);

\node[stagelabel, above=1.5mm of g-inputs.north, text=pteal!70!black] {Inputs};
\node[stagelabel, above=0.1cm of llm, text=pblue!80!black] {System};
\node[stagelabel, above=1.5mm of g-eval.north, text=pamber!80!black] {Evaluation};
\node[stagelabel, above=0.1cm of dashboard, text=ppurple!80!black] {Outputs};

\end{tikzpicture}%
}
\caption{LLM Evaluation Pipeline Overview. Test inputs flow through the LLM application to generate outputs, which are scored by automated metrics, LLM judges, and/or human evaluators. Aggregated scores feed dashboards and inform iterative improvements.}
\label{fig:pipeline-overview}
\end{figure}


\section{Evaluation Methods}
\label{sec:eval-methods}

This section organizes the major approaches to LLM evaluation around the failure modes they are best suited to detect. The choice of method depends on output structure: if the output is structured (JSON, code), use deterministic schema checks. If the task involves retrieval, add citation and grounding tests. If the output is open-ended prose, use rubric-based human evaluation or LLM-as-judge.

\subsection{Automated Offline Checks}

Automated checks provide the fastest feedback signal. \textbf{Assertions} verify deterministic properties: JSON validity, presence of required keywords, or exclusion of prohibited terms. \textbf{Golden Set Evaluation} compares outputs against verified reference answers using semantic similarity metrics (BERTScore) or exact matching for extraction tasks.

\textbf{Metamorphic testing} evaluates consistency without ground truth by checking that semantically equivalent inputs (e.g., paraphrased queries) yield consistent outputs. This detects brittleness even when ``correct'' answers are subjective~\citep{zhu2023promptbench}.

\subsection{Human Evaluation}

Human evaluation remains the primary reference point for subjective dimensions, although it is itself sensitive to rubric design, annotator expertise, and inter-rater agreement. \textbf{Rubric scoring} assigns absolute ratings (1--5) based on explicit criteria, while \textbf{Pairwise preference} asks evaluators to choose the better of two responses. Pairwise comparison often yields higher inter-rater agreement than absolute scoring because it simplifies the cognitive task~\citep{zheng2023judging}.

\subsection{Comparative Analysis}
\label{sec:eval-tradeoffs}

Table~\ref{tab:eval-tradeoffs} compares the utility of these methods. The ranges summarize values commonly reported across prior studies and should be read as approximate operating regimes rather than measurements from our experiment. Automated metrics are fast but correlate weakly with human judgment on open-ended tasks. Human evaluation is accurate but expensive. LLM-as-judge (Section~\ref{sec:llm-judges}) offers a middle ground for regression testing when calibrated against human labels.

\begin{table}[h]
\centering
\footnotesize
\caption{Approximate trade-offs in evaluation methods. Correlation ranges summarize literature-reported agreement with human labels and vary substantially by task.}
\label{tab:eval-tradeoffs}
\begin{tabular}{@{}p{2.5cm}cccc@{}}
\toprule
\textbf{Method} & \textbf{Correlation} & \textbf{Cost/1k} & \textbf{Time} & \textbf{Regression} \\
\midrule
Human eval & 1.0 (baseline) & \$1000+ & Days & High \\
LLM-as-judge & 0.70--0.85 & \$10--50 & Hours & Medium \\
BERTScore & 0.40--0.60 & \$0.10 & Mins & Low \\
Exact match & N/A & \$0.01 & Secs & High (Specific tasks) \\
\bottomrule
\end{tabular}
\end{table}

\subsection{Minimum Viable Evaluation Suite (MVES)}
\label{sec:mves}

We propose MVES as a minimum artifact standard for application-level LLM evaluation. Its purpose is to make evaluations inspectable: a reader should be able to identify the system contract, the failure modes covered, the metrics used, the evidence supporting those metrics, and the artifacts needed to reproduce the decision. Figure~\ref{fig:mves-overview} summarizes this structure. MVES is not intended as a universal sufficiency criterion; rather, it defines a minimum set of inspectable artifacts for comparing evaluation decisions across iterations. The thresholds below should be treated as starting points for calibration rather than universal requirements; teams should adjust them according to domain risk, traffic volume, and observed production failures.

\paragraph{MVES-Core (All Apps).} Every application should maintain a stratified golden set, explicit acceptance checks for contract violations, versioned prompts and model identifiers, and a human-reviewed calibration sample. For early-stage systems, a practical lower bound is 100 examples with deliberate coverage of edge cases and adversarial inputs; higher-risk systems require larger samples and tighter confidence intervals.

\paragraph{MVES-RAG (Retrieval-Based).} In addition to Core, RAG systems should evaluate retrieval and generation separately. Minimum artifacts include gold source-document annotations, retrieval metrics such as Recall@k or MRR, citation presence and citation-support checks, and explicit tests for answerable and unanswerable questions. A Recall@5 target near 0.8 is a heuristic starting point, not a universal threshold.

\paragraph{MVES-Agentic (Tool-Use).} Agentic systems add trajectory-level evaluation, per-tool success rates, sandboxed execution, and review of irreversible or high-stakes actions. A minimal suite should include multi-step tasks that exercise planning, tool selection, error recovery, and refusal behavior.


\begin{figure}[htbp]
\centering
\resizebox{0.95\textwidth}{!}{%
\begin{tikzpicture}[
    node distance=0.9cm,
    every node/.style={font=\small},
    mvesbox/.style={
        rectangle,
        rounded corners=2.5pt,
        minimum height=1.05cm,
        minimum width=2.5cm,
        text width=2.4cm,
        text centered,
        align=center,
        line width=0.7pt,
        font=\small\bfseries,
        drop shadow={shadow xshift=0.5mm, shadow yshift=-0.5mm,
                     fill=black, opacity=0.12},
    },
    exnote/.style={
        font=\scriptsize,
        text=pslate!90,
        text width=2.6cm,
        text centered,
        align=center,
    },
    mvesarrow/.style={
        -{Stealth[length=2.4mm, width=1.8mm]},
        line width=0.8pt,
        draw=pslate!75,
    },
]

\node[mvesbox, draw=pteal!80!black,   fill=pteal!8]   (app)   {Application\\Class};
\node[mvesbox, draw=pred!80!black,    fill=pred!8,    right=of app]   (fail)  {Failure\\Modes};
\node[mvesbox, draw=pamber!80!black,  fill=pamber!8,  right=of fail]  (tests) {Tests \&\\Metrics};
\node[mvesbox, draw=pblue!80!black,   fill=pblue!8,   right=of tests] (arts)  {Required\\Artifacts};
\node[mvesbox, draw=ppurple!80!black, fill=ppurple!8, right=of arts]  (valid) {Validation\\Evidence};

\draw[mvesarrow] (app) -- (fail);
\draw[mvesarrow] (fail) -- (tests);
\draw[mvesarrow] (tests) -- (arts);
\draw[mvesarrow] (arts) -- (valid);

\node[exnote, below=2mm of app]   {general\\RAG\\agentic};
\node[exnote, below=2mm of fail]  {format drift\\unsupported claims\\tool misuse};
\node[exnote, below=2mm of tests] {golden sets\\Recall@k, MRR\\citation checks};
\node[exnote, below=2mm of arts]  {versioned prompts\\gold labels\\trace logs};
\node[exnote, below=2mm of valid] {confidence intervals\\human audits\\failure replay};

\end{tikzpicture}%
}
\caption{The MVES artifact chain. Each application class maps to explicit failure modes; each failure mode is covered by tests and metrics; each test depends on concrete, versioned artifacts; and each metric is backed by validation evidence. An evaluation is auditable when every component is traceable along this chain.}
\label{fig:mves-overview}
\end{figure}

\paragraph{Relation to Existing Tools.} MVES differs from benchmark suites and metric libraries in scope. Benchmark suites such as HELM~\citep{liang2023holistic} primarily compare model behavior on predefined tasks, while metric libraries and RAG evaluators such as RAGAS~\citep{es2024ragas} provide scoring functions for particular output properties. MVES instead specifies the minimum artifacts needed to audit an application-level evaluation decision: the system contract, failure modes, test cases, metrics, prompt and model versions, and validation evidence. It can incorporate tools such as RAGAS or the lm-evaluation-harness~\citep{eleutherai2023lmeval}, but it is not a replacement for them. It also differs from checklist-style guidance by requiring each evaluation component to be traceable to a failure mode, an artifact, and a validation signal.

\begin{table}[h]
\centering
\footnotesize
\caption{MVES components, covered failure modes, required artifacts, and validation evidence.}
\label{tab:mves-components}
\begin{tabular}{@{}p{2.1cm}p{2.9cm}p{3.2cm}p{3.0cm}@{}}
\toprule
\textbf{Tier} & \textbf{Failure Modes} & \textbf{Required Artifacts} & \textbf{Validation Evidence} \\
\midrule
Core & Format drift, prompt regressions, unsafe refusals, distribution gaps & Golden set, prompt/model versions, automated checks, human calibration sample & Pass-rate confidence intervals; human audit of metric alignment \\
RAG & Retrieval misses, unsupported claims, missing or incorrect citations & Gold source IDs, retrieval logs, citation checks, unanswerable cases & Recall@k/MRR; citation-support audit; faithfulness or entailment spot checks \\
Agentic & Tool misuse, invalid trajectories, unsafe actions, poor recovery & Task traces, tool logs, sandbox outputs, approval records & Per-tool success rates; trajectory review; failure replay in sandbox \\
\bottomrule
\end{tabular}
\end{table}

Table~\ref{tab:mves-components} makes this binding concrete: each tier pairs failure modes with the artifacts that detect them and the evidence that validates the detection. Without all three elements, an evaluation may be useful for debugging but is difficult to audit or compare across iterations.


\section{Experimental Demonstration: Evaluation-Driven Iteration}
\label{sec:experiments}

This section reports a reproducible local case study of evaluation-driven prompt iteration. The ablation is intended as a reproducible demonstration of prompt-regression behavior, not as a comparative benchmark of model quality. The experiment asks a narrow mechanism question: when a prompt change appears to be a generic improvement, which component changes task-specific acceptance rates?

\subsection{Experimental Setup}

All experiments were run locally with Ollama API version 0.18.2 on a Mac mini M4 with 16GB unified memory, macOS 26.5, arm64. The expanded runs saved in the repository use the model tags \texttt{llama3:8b-instruct-q4\_K\_M} and \texttt{qwen2.5:7b-instruct-q4\_K\_M}. Decoding used temperature $=0$ and \texttt{num\_predict=1024}. Each expanded condition was run once per case; the harness also supports repeated runs and records exact-output stability. Failed model calls are recorded as failed cases with the error message preserved in the JSON output. Raw JSON validity requires the full response to parse as JSON; extraction from surrounding prose is disabled unless a test explicitly opts into it.

The expanded suites contain 30 cases per task. Each suite begins with the hand-authored seed cases and then adds deterministic synthetic variants from version-controlled templates. The \textbf{Extraction} suite covers contact information, invoice parsing, calendar events, support tickets, classification, numeric extraction, policies, meeting notes, and edge cases. The \textbf{RAG} suite contains closed-context question-answering cases where responses are checked for citation markers and expected source-derived answer terms. The \textbf{Instruction} suite covers format constraints, exact matching, refusal behavior, structured output, and pattern matching.

\subsection{Ablation Design}

The ablation is the central empirical test. We compare five prompt conditions:
\begin{itemize}[itemsep=2pt]
    \item \textbf{A (Baseline)}: task-specific prompt plus minimal system prompt.
    \item \textbf{B (+Wrapper)}: baseline prompt plus a short instruction-following wrapper.
    \item \textbf{C (+Rules)}: baseline prompt plus generic rules appended to the user prompt.
    \item \textbf{D (Full Improved)}: baseline prompt plus the full improved system prompt.
    \item \textbf{E (Non-conflicting)}: baseline prompt plus general guidance that explicitly preserves JSON-only, exact-format, and source-only task contracts.
\end{itemize}

Condition C tests whether generic rules appended to the task can change behavior. Condition D tests the full prompt-engineering intervention. Condition E tests whether regressions require direct instruction conflict or can also arise from additional contract-preserving guidance. The expectation is not that one condition should dominate everywhere; the hypothesis is that effects will vary by task contract.

\subsection{Metrics}

The harness reports \textbf{all-pass rate}, the percentage of cases for which every configured check passes, and \textbf{check-pass rate}, the percentage of individual checks that pass across cases. All checks are simple, deterministic functions: raw JSON parsing, required-key validation, allowed-value checks, citation-marker detection, expected answer-term inclusion, bullet and sentence counts, word counts, exact matching, regular-expression matching, and refusal phrase detection.

Two qualifications are important. First, JSON validity is treated as a strict response contract: the response itself must parse as JSON unless a test explicitly opts into extraction from surrounding prose. Second, the RAG checks are citation/content-compliance proxies. They do not prove claim-level faithfulness; that stronger claim would require human annotation, entailment modeling, or a validated LLM-as-judge protocol.

\subsection{Expanded Results}

Table~\ref{tab:expanded-ablation} reports all-pass counts, pass rates, and Wilson 95\% confidence intervals. Counts are shown because the suites remain modest: each 3.3 percentage-point change corresponds to one case out of 30. Figure~\ref{fig:ablation-bars} visualizes the same results and makes the task-dependent direction of each intervention immediately visible.

\begin{table}[h]
\centering
\tiny
\caption{Expanded ablation all-pass results. Cells show passed/total, pass rate, and Wilson 95\% CI.}
\label{tab:expanded-ablation}
\resizebox{\textwidth}{!}{%
\begin{tabular}{@{}llccccc@{}}
\toprule
\textbf{Model} & \textbf{Task} & \textbf{A Base} & \textbf{B Wrap} & \textbf{C Rules} & \textbf{D Full} & \textbf{E Non-conf.} \\
\midrule
Llama 3 & Extraction & 0/30, 0.0 [0.0,11.4] & 0/30, 0.0 [0.0,11.4] & 2/30, 6.7 [1.8,21.3] & 28/30, 93.3 [78.7,98.2] & 29/30, 96.7 [83.3,99.4] \\
Llama 3 & RAG & 19/30, 63.3 [45.5,78.1] & 19/30, 63.3 [45.5,78.1] & 22/30, 73.3 [55.6,85.8] & 16/30, 53.3 [36.1,69.8] & 16/30, 53.3 [36.1,69.8] \\
Llama 3 & Instruction & 20/30, 66.7 [48.8,80.8] & 20/30, 66.7 [48.8,80.8] & 20/30, 66.7 [48.8,80.8] & 24/30, 80.0 [62.7,90.5] & 24/30, 80.0 [62.7,90.5] \\
\addlinespace
Qwen 2.5 & Extraction & 4/30, 13.3 [5.3,29.7] & 4/30, 13.3 [5.3,29.7] & 22/30, 73.3 [55.6,85.8] & 30/30, 100.0 [88.6,100.0] & 25/30, 83.3 [66.4,92.7] \\
Qwen 2.5 & RAG & 26/30, 86.7 [70.3,94.7] & 24/30, 80.0 [62.7,90.5] & 9/30, 30.0 [16.7,47.9] & 16/30, 53.3 [36.1,69.8] & 21/30, 70.0 [52.1,83.3] \\
Qwen 2.5 & Instruction & 20/30, 66.7 [48.8,80.8] & 20/30, 66.7 [48.8,80.8] & 17/30, 56.7 [39.2,72.6] & 19/30, 63.3 [45.5,78.1] & 23/30, 76.7 [59.1,88.2] \\
\bottomrule
\end{tabular}
}
\end{table}


\begin{figure}[htbp]
\centering
\begin{tikzpicture}
\begin{groupplot}[
    group style={
        group size=2 by 1,
        horizontal sep=1.1cm,
        ylabels at=edge left,
    },
    ybar,
    /pgf/bar width=4.5pt,
    width=0.52\textwidth,
    height=5.6cm,
    ymin=0, ymax=112,
    ytick={0,20,40,60,80,100},
    ylabel={All-pass rate (\%)},
    ylabel near ticks,
    symbolic x coords={Extraction, RAG, Instruction},
    xtick=data,
    enlarge x limits=0.22,
    tick label style={font=\footnotesize},
    label style={font=\footnotesize},
    title style={font=\small\bfseries},
    legend columns=5,
    legend style={
        font=\footnotesize,
        at={(1.1,1.22)},
        anchor=south,
        /tikz/every even column/.append style={column sep=0.35cm},
        draw=gray!50,
    },
    legend image code/.code={\draw[#1] (0cm,-0.1cm) rectangle (0.18cm,0.14cm);},
    error bars/y dir=both,
    error bars/y explicit,
    error bars/error bar style={pslate!70, line width=0.5pt},
    error bars/error mark options={pslate!70, rotate=90, mark size=1.1pt},
    every axis plot/.append style={draw opacity=0.85},
]
\nextgroupplot[title={Llama 3 8B Instruct (q4\_K\_M)}]
\addplot+[fill=pslate!35, draw=pslate!85!black] coordinates {
    (Extraction,0.0)    +- (0,11.4) -= (0,0.0)
    (RAG,63.3)          +- (0,14.8) -= (0,17.8)
    (Instruction,66.7)  +- (0,14.1) -= (0,17.9)
};
\addplot+[fill=pblue!45, draw=pblue!85!black] coordinates {
    (Extraction,0.0)    +- (0,11.4) -= (0,0.0)
    (RAG,63.3)          +- (0,14.8) -= (0,17.8)
    (Instruction,66.7)  +- (0,14.1) -= (0,17.9)
};
\addplot+[fill=pamber!55, draw=pamber!85!black] coordinates {
    (Extraction,6.7)    +- (0,14.6) -= (0,4.9)
    (RAG,73.3)          +- (0,12.5) -= (0,17.7)
    (Instruction,66.7)  +- (0,14.1) -= (0,17.9)
};
\addplot+[fill=pred!50, draw=pred!85!black] coordinates {
    (Extraction,93.3)   +- (0,4.9)  -= (0,14.6)
    (RAG,53.3)          +- (0,16.5) -= (0,17.2)
    (Instruction,80.0)  +- (0,10.5) -= (0,17.3)
};
\addplot+[fill=pgreen!45, draw=pgreen!85!black] coordinates {
    (Extraction,96.7)   +- (0,2.7)  -= (0,13.4)
    (RAG,53.3)          +- (0,16.5) -= (0,17.2)
    (Instruction,80.0)  +- (0,10.5) -= (0,17.3)
};
\legend{A Baseline, B +Wrapper, C +Rules, D Full, E Non-conf.}

\nextgroupplot[title={Qwen 2.5 7B Instruct (q4\_K\_M)}]
\addplot+[fill=pslate!35, draw=pslate!85!black] coordinates {
    (Extraction,13.3)   +- (0,16.4) -= (0,8.0)
    (RAG,86.7)          +- (0,8.0)  -= (0,16.4)
    (Instruction,66.7)  +- (0,14.1) -= (0,17.9)
};
\addplot+[fill=pblue!45, draw=pblue!85!black] coordinates {
    (Extraction,13.3)   +- (0,16.4) -= (0,8.0)
    (RAG,80.0)          +- (0,10.5) -= (0,17.3)
    (Instruction,66.7)  +- (0,14.1) -= (0,17.9)
};
\addplot+[fill=pamber!55, draw=pamber!85!black] coordinates {
    (Extraction,73.3)   +- (0,12.5) -= (0,17.7)
    (RAG,30.0)          +- (0,17.9) -= (0,13.3)
    (Instruction,56.7)  +- (0,15.9) -= (0,17.5)
};
\addplot+[fill=pred!50, draw=pred!85!black] coordinates {
    (Extraction,100.0)  +- (0,0.0)  -= (0,11.4)
    (RAG,53.3)          +- (0,16.5) -= (0,17.2)
    (Instruction,63.3)  +- (0,14.8) -= (0,17.8)
};
\addplot+[fill=pgreen!45, draw=pgreen!85!black] coordinates {
    (Extraction,83.3)   +- (0,9.4)  -= (0,16.9)
    (RAG,70.0)          +- (0,13.3) -= (0,17.9)
    (Instruction,76.7)  +- (0,11.5) -= (0,17.6)
};
\end{groupplot}
\end{tikzpicture}
\caption{Expanded ablation all-pass rates by prompt condition (30 cases per task; error bars are Wilson 95\% confidence intervals). The same prompt interventions move tasks in opposite directions: output-contract prompts (D, E) lift strict extraction from near zero while RAG citation/content-compliance peaks under different conditions per model, and generic user-prompt rules (C) collapse Qwen 2.5 RAG compliance from 86.7\% to 30.0\%.}
\label{fig:ablation-bars}
\end{figure}

The results support three conclusions. First, the effect of a generic prompt addition is not monotonic. Stronger output-contract prompts substantially improved strict extraction in this local stack, especially because the minimal baseline frequently produced non-raw JSON. This extraction result should therefore be read as evidence that minimal baselines are inadequate for strict raw-JSON contracts, not as evidence that any generic prompt is intrinsically superior. Second, RAG citation/content-compliance is more fragile: Qwen 2.5 dropped from 26/30 in the baseline to 9/30 under generic user-prompt rules, and Llama 3 dropped from 19/30 to 16/30 under the full and non-conflicting system prompts. Third, the non-conflicting condition is informative but not uniformly protective: it improved or preserved some instruction and extraction checks, but it still reduced Llama 3 RAG all-pass relative to baseline.

\subsection{Check-Pass Results}

The all-pass metric counts a case as failed if any single check fails. Table~\ref{tab:expanded-checkpass} therefore reports check-pass rates, which distinguish conditions that fail most checks from conditions that miss one component of a case.

\begin{table}[h]
\centering
\footnotesize
\caption{Expanded ablation check-pass rates (\%).}
\label{tab:expanded-checkpass}
\begin{tabular}{@{}llccccc@{}}
\toprule
\textbf{Model} & \textbf{Task} & \textbf{A} & \textbf{B} & \textbf{C} & \textbf{D} & \textbf{E} \\
\midrule
Llama 3 & Extraction & 0.0 & 0.0 & 11.8 & 94.1 & 97.1 \\
Llama 3 & RAG & 74.4 & 74.4 & 81.4 & 65.1 & 67.4 \\
Llama 3 & Instruction & 69.2 & 71.2 & 69.2 & 78.8 & 80.8 \\
Qwen 2.5 & Extraction & 23.5 & 23.5 & 76.5 & 100.0 & 85.3 \\
Qwen 2.5 & RAG & 90.7 & 86.0 & 51.2 & 66.3 & 79.1 \\
Qwen 2.5 & Instruction & 71.2 & 71.2 & 63.5 & 69.2 & 80.8 \\
\bottomrule
\end{tabular}
\end{table}

\subsection{Failure Analysis}

The failures were inspected using saved response previews and per-check details. In extraction, the dominant failure mode for the minimal baseline was raw JSON contract violation: outputs often contained markdown fences, explanations, or malformed structures. In RAG, automated failures were missing citation markers, missing expected source-derived terms, or answers that did not use the expected unknown-answer policy. In instruction-following, failures included wrong output counts, regex mismatches, and insufficient refusals.

\begin{table}[h]
\centering
\caption{Representative failure categories exposed by the expanded ablation.}
\label{tab:failure-breakdown}
\begin{tabular}{@{}p{2.7cm}p{6.5cm}p{2.8cm}@{}}
\toprule
\textbf{Task} & \textbf{Failure Type} & \textbf{Primary Signal} \\
\midrule
Extraction & Raw JSON contract violation, markdown wrapping, or missing required structure & \texttt{json\_valid}, \texttt{required\_keys} \\
RAG & Citation-format or expected answer-term proxy failure & \texttt{must\_cite}, \texttt{gold\_contains} \\
Instruction & Wrong counts, exact-match failures, regex failures, or weak refusal behavior & Format/refusal checks \\
\bottomrule
\end{tabular}
\end{table}

These failures illustrate the operational value of case-level checks. A prompt that appears more explicit can still shift behavior away from a product contract, especially when the contract prioritizes terse structured output or evidence-bounded answers.

\subsection{Reproducibility Boundary}

The saved expanded runs fix the model tags, quantization, operating system, hardware class, decoding temperature, output length, prompt conditions, deterministic synthetic expansions, parsing rules, and failure logging. They do not fix every possible source of variation. Ollama version, model build, Metal runtime behavior, hardware state, and future model updates can change outputs. Because the expanded run uses one live generation per case and condition, it should not be read as a variance estimate. Repeated-run modes exist in the harness, but broader determinism claims should be validated separately for any new stack.

\subsection{Threats to Validity}

Several limitations constrain the scope of these results.

\paragraph{Small synthetic test suites.} The expanded suites contain 30 cases per task, not hundreds. The results are useful as a mechanism-oriented case study, but they are not sufficient for estimating production reliability.

\paragraph{Synthetic augmentation.} The added cases are deterministic and auditable, but they are template-derived. They expand coverage of schemas and constraints; they do not substitute for production-distribution sampling.

\paragraph{Prompt strawman risk.} The baseline and generic prompts are simple by construction. A stronger hand-tuned baseline might reduce or reverse some observed differences. The non-conflicting condition is included as a mitigation, but it does not exhaust the design space.

\paragraph{Proxy RAG metrics.} The RAG suite checks citation markers and expected answer terms. It does not establish citation precision, citation recall, or full claim-level entailment. The paper therefore treats RAG results as citation/content-compliance proxies, not as definitive faithfulness measurements.

\paragraph{Limited model coverage.} The experiments cover two local 7--8B quantized models. Larger open-weight models and hosted API models may exhibit different sensitivity to the same prompt changes.

\paragraph{Single hardware and inference stack.} Local Ollama runs improve reproducibility and privacy, but results may vary across quantization schemes, Ollama versions, operating systems, and hardware.

\paragraph{Author-designed prompts and tests.} The same author designed the tasks, prompts, and checks, which can introduce unintentional bias. A stronger validation would use independently authored prompts and held-out production traces.

These limitations motivate a conservative interpretation: the experiments show that prompt changes can produce task-specific trade-offs, and that a lightweight harness can expose these trade-offs before deployment. They do not establish universal prompt-engineering laws.

\begin{keytakeaways}
Prompt changes should be evaluated against task-specific acceptance criteria. In this expanded local demonstration, stronger output-contract prompts helped strict extraction, but RAG citation/content-compliance regressed under several generic-rule conditions. The result is not ``generic prompts are bad''; it is that prompt changes are empirical interventions and should be regression tested.
\end{keytakeaways}

\subsection{Reproducibility}

The evaluation code is in \texttt{eval\_harness/}. The main runner is \texttt{run\_eval.py}; the ablation runner is \texttt{run\_ablation.py}. Both write timestamped JSON outputs and generated tables under \texttt{eval\_harness/results/} or \texttt{eval\_harness/results\_ablation/}. Dry-run modes are provided to test the pipeline without invoking Ollama.

To reproduce the expanded ablation, install the listed Python dependencies, start Ollama, pull the model tags reported above, and run:
\begin{lstlisting}[basicstyle=\small\ttfamily]
python run_ablation.py --dataset all --augment-size 30 --runs 1 \
  --model llama3:8b-instruct-q4_K_M
python run_ablation.py --dataset all --augment-size 30 --runs 1 \
  --model qwen2.5:7b-instruct-q4_K_M
\end{lstlisting}
To run a larger but slower experiment on comparable hardware, increase \texttt{--augment-size} and \texttt{--runs}; the 16GB Mac mini configuration should run models sequentially rather than in parallel.


\section{Quality Taxonomy}
\label{sec:quality-taxonomy}

Before designing evaluations, we must articulate what ``quality'' means for a given application. This section defines the quality dimensions that MVES (Section~\ref{sec:mves}) uses to map application requirements to measurable failure modes. Different applications weight these dimensions differently, so teams should identify which dimensions matter most before investing in evaluation infrastructure.

\subsection{Correctness}

Correctness measures whether the LLM's output is factually accurate and logically sound. This dimension matters most for applications involving factual claims, calculations, or procedural instructions. An output is considered correct if it accurately reflects ground truth, follows valid reasoning, and contains no factual errors.

Correctness evaluation requires reference answers or verifiable facts. For question-answering tasks, this may involve comparing outputs against gold-standard answers. For reasoning tasks, evaluators must verify the logical chain. In practice, correctness is often the most tractable dimension to evaluate because it admits objective verification.

\subsection{Helpfulness}

Helpfulness measures whether the output actually assists the user in achieving their goal. An answer may be technically correct but unhelpful if it is incomplete, overly verbose, or misinterprets the user's intent. A helpful output addresses the user's underlying intent, provides actionable information, and is appropriately scoped to the question asked.

Helpfulness is often subjective and context-dependent, making it well-suited for human evaluation or preference-based comparisons~\citep{ouyang2022training}. The RLHF training paradigm explicitly optimizes for human preferences on helpfulness, demonstrating the centrality of this dimension to modern LLM development.

\subsection{Harmlessness}

Harmlessness measures whether the output avoids causing harm through dangerous advice, toxic content, privacy violations, or manipulation. A harmless output does not promote violence, contain hate speech, reveal private information, or provide instructions for dangerous activities.

The harmlessness dimension has received significant attention in the alignment literature. Constitutional AI~\citep{bai2022constitutional} provides frameworks for encoding safety constraints directly into model training. Evaluation for harmlessness typically involves adversarial testing with prompts designed to elicit harmful responses, combined with human review of edge cases.

\subsection{Groundedness and Attribution}

Groundedness measures whether claims made by the LLM can be traced to reliable sources. This dimension is especially important for RAG systems and applications where users expect cited evidence. An output is grounded if every factual claim can be attributed to a source in the provided context or a verifiable external reference.

A response may be correct but ungrounded, meaning the claim is true but not supported by provided sources. Conversely, a response may be grounded but incorrect if the source itself is wrong or misinterpreted. The distinction is central when evaluating retrieval-augmented systems~\citep{min2023factscore}. Users of knowledge-intensive applications often care more about verifiability than mere correctness.

\subsection{Refusal Correctness}

LLMs are often designed to refuse certain requests, including those that are harmful, outside scope, or unanswerable given available information. Refusal correctness measures whether the model refuses appropriately. This dimension has two components: correct refusals (the model refuses requests it should refuse) and incorrect refusals, also called over-refusal (the model refuses benign requests it should answer).

Over-refusal degrades user experience and can undermine trust. Evaluation should measure both false negatives (failure to refuse harmful requests) and false positives (refusing harmless requests). Finding the right balance requires careful test set design with both harmful prompts that should be refused and edge cases that appear problematic but are actually benign.

\subsection{Format and Style Adherence}

Many applications require outputs in specific formats such as JSON, Markdown, or particular tones or length constraints. Format adherence measures compliance with these structural requirements. An output demonstrates format adherence if it matches the specified structure, syntax, tone, and length constraints.

Format violations may cause downstream parsing failures in programmatic applications or user dissatisfaction in conversational ones. This dimension is often tested with automated validators that check structural compliance before semantic evaluation begins.

\subsection{Consistency}

Consistency measures whether the model provides coherent answers across related queries and maintains positions stated earlier in a conversation. An output is consistent if it does not contradict itself, prior model statements in the conversation, or known facts about the domain.

Evaluating consistency often requires multi-turn evaluation or metamorphic testing approaches. Inconsistency can be particularly problematic in conversational applications where users may ask follow-up questions that probe previously stated positions.

\subsection{Mapping Dimensions to Applications}

Different applications weight these dimensions differently. Table~\ref{tab:quality-dimensions} provides guidance on dimension importance across common application types.

\begin{table}[h]
\centering
\caption{Quality dimension importance by application type.}
\label{tab:quality-dimensions}
\begin{tabular}{@{}lccccc@{}}
\toprule
\textbf{Application} & \textbf{Correct.} & \textbf{Helpful.} & \textbf{Harmless.} & \textbf{Grounded.} & \textbf{Format} \\
\midrule
Customer support   & Med  & High & High & Med  & Med \\
Medical Q\&A       & High & High & High & High & Med \\
Code generation    & High & High & Med  & Med  & High \\
Creative writing   & Low  & High & Med  & Low  & Med \\
RAG knowledge base & High & Med  & Med  & High & Med \\
\bottomrule
\end{tabular}
\end{table}


\section{Test Set Design}
\label{sec:testset-design}

The quality of an evaluation depends heavily on the test set; in MVES terms, the test set is the core required artifact, encoding the failure modes a team has decided to detect. This section provides guidance on constructing test sets that meaningfully assess LLM application quality. Table~\ref{tab:dataset-summary} summarizes the hand-authored seed suites and the deterministic 30-case expanded suites used in the reported ablation.

\begin{table}[h]
\centering
\caption{Test suite summary: what each dataset evaluates.}
\label{tab:dataset-summary}
\begin{tabular}{@{}lp{4cm}p{4cm}@{}}
\toprule
\textbf{Dataset} & \textbf{What It Tests} & \textbf{Failure Caught} \\
\midrule
Extraction (20 seed / 30 expanded) & JSON schema + required keys & Format drift, markdown wrappers \\
RAG (15 seed / 30 expanded) & Citation markers + expected source-derived terms & Missing citations, answer-term gaps \\
Instruction (15 seed / 30 expanded) & Format constraints + refusal behavior & Wrong counts, noncompliance \\
\bottomrule
\end{tabular}
\end{table}

\subsection{Representativeness}

A test set should mirror the distribution of inputs the system will encounter in production. Achieving representativeness requires deliberate sampling strategies. Production sampling collects anonymized queries from actual users when available, providing direct evidence of real usage patterns. Task decomposition enumerates the categories of requests the system should handle and samples from each category. Stakeholder input from subject-matter experts can identify common and critical use cases that might not appear in logs.

Stratification ensures test cases cover different input types and lengths, various user intents, multiple difficulty levels, and all supported languages or domains. Without stratification, test sets may over-represent easy cases or common queries while under-representing the long tail of inputs that often cause production failures.

\subsection{Edge Cases and Adversarial Prompts}

Beyond typical inputs, test sets should include challenging cases that stress the system. Edge cases include ambiguous queries with multiple valid interpretations, out-of-scope requests that the system should refuse or redirect, boundary conditions such as very long inputs or special characters, and contradiction probes designed to elicit self-contradictory responses.

Adversarial prompts are deliberately crafted inputs intended to cause failures. These include prompt injection attempts that try to override system instructions, jailbreaking prompts that attempt to bypass safety guardrails, inputs designed to trigger hallucination, and format-breaking inputs that attempt to corrupt structured outputs. Red-teaming efforts~\citep{perez2022red, ganguli2022red} have codified methodologies for systematic adversarial testing.

\subsection{Systematic Coverage Design}

Achieving broad coverage requires systematic strategies beyond random sampling.

\paragraph{Intent Stratification.} Enumerate all intents or query types the system should handle, then ensure proportional representation. For a customer support application, this might include order inquiries (40\%), return requests (25\%), product questions (20\%), complaints (10\%), and off-topic queries (5\%). Document the target distribution and measure actual coverage.

\paragraph{Difficulty Stratification.} Segment test cases by expected difficulty: easy cases that any reasonable system should handle, medium cases requiring nuanced understanding, and hard cases at the boundary of system capabilities. A common distribution targets 50\% easy, 30\% medium, and 20\% hard cases.

\paragraph{Hard Negative Mining.} Hard negatives are inputs that are similar to positive cases but should produce different outputs. For a RAG system, this includes queries that nearly match a knowledge base entry but require a different answer. Hard negatives reveal overfitting to surface patterns.

\begin{lstlisting}[basicstyle=\small\ttfamily,language=Python]
# Example: Mining hard negatives for a product FAQ
positive = "How do I return a damaged item?"  # -> Return policy
hard_neg = "How do I return an item I changed my mind about?"  
# -> Different policy, tests nuanced understanding
\end{lstlisting}

\paragraph{Failure-Driven Augmentation.} When production failures occur, systematically add similar cases to the test set. This creates a living test suite that captures the failure modes discovered over time. Track the provenance of each test case (synthetic, production sample, or failure-derived).

\subsection{Illustrative Example: The Extraction Evaluation Loop}

To illustrate the cycle for structured data tasks:

\begin{enumerate}
    \item \textbf{Goal}: Extract valid JSON objects for API consumption.
    \item \textbf{Test}: Run a golden set of 20 invoices with varied formats.
    \item \textbf{Failure}: A generic helpful prompt (``Extract the information'') produces:
    \begin{badexample}
    Sure, here is the data:\\
    ```json\\
    \{ "total": "\$500" \}\\
    ```\\
    (Fails validation due to markdown blocks and conversational filler)
    \end{badexample}
    \item \textbf{Fix}: Switch to a task-specific constraint prompt: ``Output VALID JSON ONLY. Do not include markdown formatting or conversational text.''
    \item \textbf{Re-test}: The targeted prompt yields clean, parseable JSON:
    \begin{goodexample}
    \{ "total": 500.00 \}
    \end{goodexample}
\end{enumerate}

This highlights the finding from Section~\ref{sec:experiments} that task-specific constraints can outperform generic helpfulness in the local case study.

\subsection{Multi-Turn Conversation Tests}

Many LLM applications involve multi-turn dialogue. Evaluation must assess behavior across conversation trajectories, not just single-turn performance. Key considerations include context retention (whether the model correctly references earlier turns), consistency (whether the model contradicts itself across turns), clarification handling (whether the model responds appropriately to follow-up questions), and topic switching (how the model handles abrupt topic changes).

Multi-turn test cases should specify the full conversation history, expected behaviors at each turn, and evaluation criteria. This format is more complex to author than single-turn cases but essential for applications where conversation quality matters.

\subsection{Data Contamination Considerations}

Modern LLMs are trained on massive web corpora that may include common benchmarks. If evaluation test cases appear in training data, performance estimates are inflated and unreliable.

\begin{warningbox}
Data contamination is a growing concern as training corpora expand. Never assume that a public benchmark provides uncontaminated evaluation.
\end{warningbox}

Mitigation strategies include creating proprietary test sets using internal data not available on the web, date filtering to use content created after the model's training cutoff, perturbation testing to paraphrase test cases and verify consistent performance, and contamination detection to test whether the model can recite test cases verbatim~\citep{sainz2023nlp, jacovi2023stop}.

\subsection{Test Set Size and Statistical Power}

Determining the appropriate number of test cases requires considering several factors. Effect size matters: smaller expected differences require more samples to detect reliably. Variance matters: higher output variance requires more samples to achieve stable estimates. Strata matter: each category in a stratified test set needs sufficient representation to draw conclusions.

A rough guideline suggests that detecting a 5\% absolute difference in pass rate with 95\% confidence and 80\% power requires approximately 400 to 600 test cases per condition. Smaller test sets may suffice for detecting larger differences or for preliminary evaluation during development. Confidence intervals should always be reported alongside point estimates to communicate the uncertainty in measurements.

\subsection{Test Set Maintenance}

Test sets require ongoing maintenance to remain useful. Version control should track all changes to test cases, enabling reproducibility and debugging when metrics change unexpectedly. Periodic refresh adds new cases as the application evolves and usage patterns shift. Decontamination rotates out cases that may have become contaminated through inclusion in model training data. Gold answer review periodically verifies that reference answers remain accurate, especially for time-sensitive information.


\section{Metrics and Scoring}
\label{sec:metrics-scoring}

Choosing appropriate metrics is critical for meaningful evaluation; within MVES, each metric supplies the validation signal attached to a failure mode. While exact match works for deterministic tasks, natural language requires semantic assessment.

\subsection{Semantic Similarity}

Embedding-based metrics address the semantic gap by comparing meaning rather than surface form. \textbf{BERTScore} computes similarity between contextual embeddings, correlating better with human judgment than n-gram metrics like BLEU or ROUGE~\citep{zhang2020bertscore}. \textbf{BLEURT} is fine-tuned on human ratings to predict quality directly~\citep{sellam2020bleurt}. For retrieval, cosine similarity between sentence embeddings provides a directional signal of relevance.

\subsection{Factual Accuracy}

For fact-centric tasks, the \textbf{FActScore} methodology decomposes outputs into atomic claims and verifies each against a knowledge source~\citep{min2023factscore}. This granular approach reveals hallucinations that holistic scoring might miss. For example, a biography might be 90\% correct generally but fail on specific dates.

\begin{empirical}[FActScore Findings]
Min et al.\ found that ChatGPT achieved only 58\% factual precision on generated biographies, meaning 42\% of atomic claims were unsupported. Retrieval augmentation improved this to 66\%.
\end{empirical}

\subsection{Truthfulness and Calibration}

\textbf{Truthfulness} measures whether models avoid reproducing common misconceptions. The TruthfulQA benchmark shows that larger models can be \emph{less} truthful because they learn human misconceptions more effectively~\citep{lin2022truthfulqa}.

\textbf{Calibration} measures whether a model's confidence scores predict correctness. Well-calibrated models enable selective answering, where low-confidence responses are routed to human review. This is essential for high-stakes applications where errors are costly.

\subsection{Inter-Rater Reliability}

When using human evaluators, measuring agreement ensures rubric reliability. \textbf{Cohen's Kappa} measures agreement between two raters adjusted for chance. A Kappa $> 0.6$ is commonly treated as substantial agreement, while scores below 0.4 suggest the rubric is ambiguous and needs revision. \textbf{Krippendorff's Alpha} extends this to multiple raters and missing data~\citep{krippendorff2011content}.


\section{RAG Evaluation}
\label{sec:rag-evaluation}

Retrieval-Augmented Generation (RAG) systems combine information retrieval with LLM generation~\citep{lewis2020retrieval}. Evaluating these systems requires assessing both components and their interaction. This section details the evaluation methods behind the MVES-RAG tier, including the RAGAS framework and standard retrieval metrics.

\subsection{Decomposing RAG Evaluation}

A RAG system operates in two stages. In the retrieval stage, the system retrieves relevant documents from a corpus given a query. In the generation stage, it produces a response given the query and retrieved documents. Failures can occur in either stage or in their integration, so effective evaluation must isolate these sources to enable targeted improvement.

\subsection{The RAGAS Framework}

RAGAS (Retrieval Augmented Generation Assessment)~\citep{es2024ragas} provides a reference-free evaluation framework with four key metrics. Table~\ref{tab:ragas-metrics} summarizes these metrics.

\begin{table}[h]
\centering
\caption{The RAGAS metrics for RAG evaluation.}
\label{tab:ragas-metrics}
\begin{tabular}{@{}lp{7cm}l@{}}
\toprule
\textbf{Metric} & \textbf{Definition} & \textbf{Measures} \\
\midrule
Faithfulness & Fraction of claims in the answer that can be inferred from the context & Generation \\
\addlinespace
Answer Relevance & Semantic similarity between the answer and the question & Generation \\
\addlinespace
Context Precision & Whether the relevant context chunks are ranked higher & Retrieval \\
\addlinespace
Context Recall & Fraction of ground truth that is covered by retrieved context & Retrieval \\
\bottomrule
\end{tabular}
\end{table}

Faithfulness is calculated as the ratio of claims supported by context to total claims in the answer:
\begin{equation}
\text{Faithfulness} = \frac{\text{Number of claims supported by context}}{\text{Total claims in answer}}
\end{equation}

The following example illustrates faithfulness evaluation:

\begin{promptbox}[User Query]
\small What are the side effects of aspirin?
\end{promptbox}

\begin{responsebox}[Retrieved Context]
\small Aspirin may cause stomach irritation and bleeding. It can also increase the risk of kidney problems in high doses. Rare side effects include allergic reactions.
\end{responsebox}

\begin{responsebox}[LLM Answer]
\small Aspirin can cause stomach irritation, bleeding, and may affect kidney function at high doses. It also helps prevent heart attacks in some patients.
\end{responsebox}

\begin{table}[h]
\centering
\footnotesize
\caption{Faithfulness analysis of the above answer.}
\begin{tabular}{@{}p{6cm}cc@{}}
\toprule
\textbf{Claim in Answer} & \textbf{In Context?} & \textbf{Status} \\
\midrule
Aspirin can cause stomach irritation & \textcolor{green!60!black}{\checkmark} Yes & Supported \\
Aspirin can cause bleeding & \textcolor{green!60!black}{\checkmark} Yes & Supported \\
May affect kidney function at high doses & \textcolor{green!60!black}{\checkmark} Yes & Supported \\
Helps prevent heart attacks & \textcolor{red!60!black}{$\times$} No & \textbf{Unsupported} \\
\midrule
\textbf{Faithfulness Score} & \multicolumn{2}{c}{\textbf{3/4 = 75\%}} \\
\bottomrule
\end{tabular}
\end{table}

\begin{warningbox}
The claim about preventing heart attacks is correct but \textbf{unsupported by the context}. This is a ``correct but unfaithful'' response, which is a critical failure mode in RAG systems.
\end{warningbox}

\subsection{Retrieval Quality Metrics}

Standard information retrieval metrics apply to the retrieval component of RAG systems. Precision@k measures the fraction of top-k retrieved documents that are relevant. Recall@k measures the fraction of all relevant documents that appear in the top-k results. Mean Reciprocal Rank (MRR) averages the reciprocal ranks of the first relevant document across queries. Normalized Discounted Cumulative Gain (nDCG) accounts for graded relevance and position, giving more credit to relevant documents appearing earlier.

\begin{lstlisting}[basicstyle=\small\ttfamily,language=Python]
def precision_at_k(retrieved: list, relevant: set, k: int) -> float:
    """Compute Precision@k for retrieval evaluation."""
    top_k = retrieved[:k]
    relevant_in_top_k = sum(1 for doc in top_k if doc in relevant)
    return relevant_in_top_k / k

def reciprocal_rank(retrieved: list, relevant: set) -> float:
    """Compute reciprocal rank (for MRR calculation)."""
    for i, doc in enumerate(retrieved):
        if doc in relevant:
            return 1.0 / (i + 1)
    return 0.0
\end{lstlisting}

Relevance assessment requires labels indicating which documents are relevant for each query. These can be obtained through manual annotation by domain experts, implicit signals such as documents that users clicked, or LLM-based relevance judgments with appropriate calibration.

\subsection{The ``Correct but Unsupported'' Failure Mode}

A subtle failure occurs when the LLM generates a correct answer using its parametric knowledge rather than the retrieved documents. This is problematic because users cannot verify the answer against provided sources, the system may hallucinate when parametric knowledge is wrong, and it undermines the purpose of grounded, attributable responses.

\begin{badexample}[Correct but Unsupported]
\small
\textbf{Query:} Who wrote Romeo and Juliet? \\
\textbf{Context:} [Document about Shakespeare's biography, mentioning only Hamlet] \\
\textbf{Answer:} ``William Shakespeare wrote Romeo and Juliet.'' \\
\textbf{Problem:} Answer is correct but not supported by the provided context. The model used parametric knowledge instead of retrieved documents.
\end{badexample}

Detection involves comparing responses with and without retrieval; if answers are identical, the model may be ignoring context. Mitigation strategies include prompt engineering to emphasize grounding, fine-tuning on attribution data, and filtering responses that lack citations.

\paragraph{Before/After Prompt Comparison.} The following illustrates how explicit grounding constraints fix this failure:

\begin{badexample}[Baseline Prompt]
\small Answer the question using the provided sources.
\end{badexample}

\begin{goodexample}[Improved Prompt]
\small Answer using ONLY the provided sources. Cite each claim with [1], [2], etc. If the sources do not contain the answer, respond ``I don't know based on the provided sources.''
\end{goodexample}

With the improved prompt, the model responds: ``The sources discuss Shakespeare's biography and mention Hamlet, but do not reference Romeo and Juliet. I don't know based on the provided sources.'' This is a correct refusal that our evaluation harness catches as a pass.

\subsection{Citation Coverage and Quality}

When RAG systems include citations, evaluating them explicitly provides insight into attribution quality. Table~\ref{tab:citation-quality} defines the key metrics.

\begin{table}[h]
\centering
\caption{Citation quality metrics for RAG systems.}
\label{tab:citation-quality}
\begin{tabular}{@{}ll@{}}
\toprule
\textbf{Metric} & \textbf{Formula} \\
\midrule
Citation Density & Citations per 100 words of response \\
Citation Precision & $\frac{\text{Citations that support claims}}{\text{Total citations}}$ \\
Citation Recall & $\frac{\text{Claims with citations}}{\text{Total claims}}$ \\
Source Diversity & Number of unique sources cited \\
\bottomrule
\end{tabular}
\end{table}

Human annotators or LLM judges verify that cited passages actually support the claims made. This verification step catches cases where citations are present but do not substantiate the associated claim.

\subsection{End-to-End vs.\ Component Evaluation}

End-to-end evaluation measures final answer quality regardless of intermediate steps, reflecting user experience and providing simpler implementation. However, it makes diagnosing failures difficult. Component evaluation measures retrieval and generation separately, isolating issues and enabling targeted fixes, but may miss integration bugs. The recommended approach uses both: end-to-end evaluation ensures overall quality while component evaluation identifies where to invest improvement effort.

\subsection{Illustrative Example: The RAG Evaluation Loop}

To illustrate the evaluation loop in practice, consider a system prone to hallucinating information not present in the retrieved context.

\begin{enumerate}
    \item \textbf{Goal}: Ensure answers are strictly grounded in retrieved documents.
    \item \textbf{Test}: Run a golden set including questions where the answer is known \emph{outside} the system but absent from the retrieved context (see ``Correct but Unsupported'' above).
    \item \textbf{Failure}: The baseline prompt answers from parametric memory:
    \begin{badexample}
    \textbf{Question}: Is SSO included in the Business plan? \\
    \textbf{Retrieved}: [Business Plan features: shared workspaces, priority support.] \\
    \textbf{Output}: Yes, SSO is included. (Incorrectly using outside knowledge)
    \end{badexample}
    \item \textbf{Fix}: Update prompt to require citations and explicit refusal: ``Answer using ONLY the sources. Cite every claim like [1]. If the answer is not in the sources, say 'I don't know'.''
    \item \textbf{Re-test}: The improved prompt correctly refuses:
    \begin{goodexample}
    \textbf{Output}: I don't know based on the provided sources, as SSO is not listed in the Business Plan features [1].
    \end{goodexample}
\end{enumerate}

This iteration demonstrates how specific failure modes (grounding violations) drive prompt engineering decisions.


\section{LLM-as-Judge}
\label{sec:llm-judges}

Using LLMs to evaluate other LLM outputs (``LLM-as-judge'') is widely used as a scalable alternative to human evaluation~\citep{zheng2023judging}. This section examines when this approach works, when it fails, and provides concrete examples of effective judge prompts.

\subsection{Motivation for LLM-Based Judging}

LLM judges are useful when teams need a fast, inexpensive screening signal over many examples. Their main value is in regression testing and triage, not in replacing human evaluation. Because judge behavior can be biased or poorly calibrated, judge scores should be validated against human labels before being used to make quality claims; within MVES, calibrated judge scores count as validation evidence only alongside a human-reviewed sample.

\begin{empirical}[MT-Bench Findings~\citep{zheng2023judging}]
In evaluating chat assistants, GPT-4 as a judge achieved \textbf{over 80\% agreement} with human preferences, comparable to reported agreement between human annotators. This result supports LLM-as-judge as a calibrated screening tool for some evaluation settings; it does not establish judge reliability for all domains, rubrics, or risk levels.
\end{empirical}

LLM judges work well for relative comparisons between two outputs rather than absolute quality assessment. They excel on tasks where quality dimensions are well-defined and unambiguous. They are valuable for initial screening before human evaluation and for high-volume regression testing where human evaluation is impractical.

\subsection{Known Biases and Failure Modes}

LLM judges exhibit several documented biases that evaluators must account for. Table~\ref{tab:llm-biases} summarizes the key biases with reported or expected scale and mitigation strategies. The ranges are literature-derived or operational heuristics, not measurements from the local case study in Section~\ref{sec:experiments}.

\begin{table}[h]
\centering
\footnotesize
\caption{Documented biases in LLM-as-judge evaluation with mitigation strategies. Numeric entries are approximate literature-reported ranges or operational heuristics, not measurements from this paper.}
\label{tab:llm-biases}
\begin{tabular}{@{}p{2.2cm}p{4.5cm}p{2cm}p{3cm}@{}}
\toprule
\textbf{Bias} & \textbf{Description} & \textbf{Scale} & \textbf{Mitigation} \\
\midrule
Position bias & Systematic preference for first or second option & 5--15\% & Randomize order \\
\addlinespace
Verbosity bias & Preferring longer responses regardless of quality & 10--20\% & Length-normalize \\
\addlinespace
Self-preference & Preferring outputs from same model family & 10--25\% & Use different judge \\
\addlinespace
Style bias & Preferring confident tone even when wrong & Variable & Rubric-based scoring \\
\addlinespace
Instruction leakage & Rewarding rubric-hacking & Variable & Blind to criteria \\
\bottomrule
\end{tabular}
\end{table}

\paragraph{Position Bias.} LLMs can systematically favor the first or second position when comparing two outputs~\citep{wang2024large, zheng2023judging}. Reported effect sizes vary by model, task, and prompt format. Mitigation requires randomizing presentation order and averaging across both orderings.

\paragraph{Verbosity Bias.} Longer responses receive higher ratings regardless of actual quality~\citep{zheng2023judging}. This is particularly problematic when comparing a concise, correct answer against a verbose but less accurate one. Mitigation includes explicit rubric criteria penalizing unnecessary length or length-normalizing scores.

\paragraph{Self-Preference Bias.} LLMs prefer outputs generated by themselves or by similar models~\citep{panickssery2024llm}. When GPT-4 evaluates GPT-4 outputs against Claude outputs, it shows measurable preference for GPT-4 responses. Always use a judge model from a different family than the system being evaluated.

\paragraph{Style Bias.} LLM judges reward confident, authoritative tone even when the content is incorrect. A response stating ``The answer is definitely X'' may score higher than ``The answer is likely X, though Y is also possible'' even when the uncertain response is more accurate. Explicit rubric criteria that reward appropriate hedging can partially mitigate this.

\paragraph{Instruction Leakage (Rubric-Hacking).} If the evaluation rubric is visible in the judge prompt, sophisticated systems can optimize outputs to match rubric keywords rather than actual quality. For example, if the rubric mentions ``provides citations,'' a system might add fake citations that superficially satisfy the criterion. Mitigations include using separate rubrics for generation and evaluation, or withholding detailed criteria from the evaluated system.

\begin{warningbox}
Never use LLM-as-judge as the sole arbiter for high-stakes decisions. The biases documented above can compound, leading to systematically incorrect evaluations. Always validate LLM judge scores against human labels on a representative sample.
\end{warningbox}

\subsection{Good vs.\ Bad Judge Prompts}

The quality of evaluation depends heavily on prompt design.

\begin{badexample}[Vague Judge Prompt]
\begin{lstlisting}[basicstyle=\small\ttfamily]
Is Response A or Response B better?
Just say A or B.
\end{lstlisting}
\textbf{Problems:} No criteria defined; no reasoning required; prone to position bias.
\end{badexample}

\begin{goodexample}[Effective Judge Prompt]
\begin{lstlisting}[basicstyle=\small\ttfamily]
You are an expert evaluator. Compare the two responses below.

[Question]
{question}

[Response A]
{response_a}

[Response B]
{response_b}

Evaluate on these criteria (1-5 each):
1. Accuracy: Are all facts correct?
2. Completeness: Does it fully answer the question?
3. Clarity: Is it well-organized and easy to understand?
4. Conciseness: Is it appropriately brief without 
   unnecessary filler?

Briefly justify the score for each criterion.
Then provide scores in this JSON format:
{
  "analysis_a": "...",
  "analysis_b": "...",
  "scores_a": {"accuracy": X, "completeness": X, ...},
  "scores_b": {"accuracy": X, "completeness": X, ...},
  "winner": "A" or "B" or "tie",
  "reasoning": "..."
}
\end{lstlisting}
\textbf{Strengths:} explicit criteria, structured scoring, auditable justification fields, and reduced ambiguity in the judgment task.
\end{goodexample}

\subsection{Implementation Example}

The following implementation demonstrates position bias mitigation through order randomization:

\begin{lstlisting}[basicstyle=\small\ttfamily,language=Python]
import json
import random

def llm_judge_pairwise(
    question: str,
    response_a: str, 
    response_b: str,
    judge_model: str = "gpt-4"
) -> dict:
    """Evaluate two responses using an LLM judge."""
    
    # Randomize order to mitigate position bias
    if random.random() < 0.5:
        first, second = response_a, response_b
        order = "original"
    else:
        first, second = response_b, response_a
        order = "swapped"
    
    prompt = f"""
    You are an expert evaluator. 
    
    [Question] {question}
    [Response 1] {first}
    [Response 2] {second}
    
    Which response is better? Justify briefly against
    the criteria, then output JSON with your verdict.
    """
    
    result = call_llm(judge_model, prompt)
    verdict = json.loads(result)
    
    # Correct for order swapping
    if order == "swapped":
        verdict["winner"] = swap_winner(verdict["winner"])
    
    return verdict
\end{lstlisting}

\subsection{Multi-Judge Aggregation}

Using multiple judge models improves reliability by reducing the impact of model-specific biases:

\begin{lstlisting}[basicstyle=\small\ttfamily,language=Python]
def multi_judge_evaluation(question, response_a, response_b):
    """Use multiple judges and aggregate results."""
    judges = ["gpt-4", "claude-3-opus", "gemini-pro"]
    verdicts = []
    
    for judge in judges:
        verdict = llm_judge_pairwise(
            question, response_a, response_b, judge
        )
        verdicts.append(verdict["winner"])
    
    # Majority vote
    from collections import Counter
    vote_counts = Counter(verdicts)
    winner = vote_counts.most_common(1)[0][0]
    
    # Flag disagreements for human review
    agreement = vote_counts[winner] / len(judges)
    needs_human_review = agreement < 0.67
    
    return {
        "winner": winner,
        "agreement": agreement,
        "needs_human_review": needs_human_review
    }
\end{lstlisting}

\subsection{Comparison: LLM Judges vs.\ Human Evaluation}

Table~\ref{tab:llm-vs-human} compares the trade-offs between LLM judges and human evaluation.

\begin{table}[h]
\centering
\caption{LLM judges vs.\ human evaluation trade-offs.}
\label{tab:llm-vs-human}
\begin{tabular}{@{}lcc@{}}
\toprule
\textbf{Factor} & \textbf{LLM Judge} & \textbf{Human} \\
\midrule
Scale               & High       & Low \\
Cost                & Low        & High \\
Speed               & Fast       & Slow \\
Specialized domains & Variable   & High (experts) \\
Bias awareness      & Limited    & High \\
Subjective nuance   & Moderate   & High \\
Explainability      & Moderate   & High \\
\bottomrule
\end{tabular}
\end{table}

\subsection{Protocol for Robust Judge Evaluation}

To mitigate the biases documented above, we recommend a standardized mitigation protocol rather than ad-hoc prompting.

\begin{itemize}
    \item \textbf{Position Bias Mitigation.} Run every pairwise comparison twice, swapping the order (A vs B, then B vs A). If the judge's preference flips with the order, declare a tie or send the case to human review. This double-pass check exposes order-sensitive judgments directly in the evaluated sample.
    
    \item \textbf{Length Normalization.} To combat verbosity bias, instruct the judge to penalize unnecessary length, or truncate responses to the length of the shorter answer plus 20\% before judging.
    
    \item \textbf{Rubric-Conditioned Scoring.} Instead of asking ``Which is better?'', ask ``Which response better satisfies criteria X, Y, and Z?'' Component-level scoring reduces the impact of confident but incorrect answers.
    
    \item \textbf{Reference-Guided Grading.} Providing a reference answer grounds the evaluation in the case's ground truth rather than in plausibility, which reduces reliance on the judge's own knowledge.
\end{itemize}

\subsection{Recommended Guardrails}

To deploy LLM judges safely, follow a strict protocol. Never use self-evaluation; a model should not judge its own inputs. Validate judge scores against human labels on a representative sample, and predefine acceptable agreement or correlation thresholds for the target domain (for example, a correlation above 0.7 when continuous scores are appropriate). Use judges primarily for screening or regression testing, not as the final certifier for high-stakes decisions. Finally, re-validate calibration whenever the provider updates the model, as judge behavior can drift.


\section{Illustrative Application Mappings}
\label{sec:case-studies}

This section illustrates how MVES applies to three representative LLM applications. The cases are illustrative composite scenarios synthesized from common deployment patterns. They demonstrate how the proposed evaluation structure maps quality dimensions, failure modes, artifacts, and validation evidence for different application classes; they should not be read as empirical deployment studies. Each scenario describes the quality dimensions that would matter most, how the evaluation could be structured, and what failure modes the setup is intended to expose.

\subsection{Mapping 1: Customer Support Assistant}

Consider a composite e-commerce support assistant designed to handle tier-1 customer inquiries, including questions about orders, returns, shipping policies, and product availability. The system must resolve common issues without human intervention while correctly escalating complex cases to human agents.

The key quality dimensions for this application are helpfulness (whether the response actually resolves the customer's issue), harmlessness (avoidance of inappropriate content or false promises about policies), format adherence (consistency with brand voice guidelines), and escalation accuracy (correct identification of cases requiring human agents). Escalation accuracy is particularly critical: incorrect escalations either overwhelm human agents with trivial requests or leave frustrated customers without recourse.

In this scenario, the test set is constructed by sampling from historical support tickets across product categories. The team ensures coverage of edge cases including multi-issue tickets where customers raise several concerns simultaneously, emotionally charged messages from frustrated customers, and queries about policy exceptions not covered in standard documentation. An adversarial component includes prompts attempting to extract confidential information such as customer data or internal pricing rules.

The evaluation combines multiple approaches. Automated checks validate format compliance and flag responses containing prohibited phrases or policy violations. An LLM-as-judge system uses a rubric-based prompt to score helpfulness and tone on a 1-to-5 scale. A sample of 200 cases per week is reviewed by support team leads to ensure quality and identify new failure patterns. In production, the team tracks customer satisfaction surveys, resolution rates, and escalation rates as ground-truth signals.

The scenario highlights several practical lessons. Escalation accuracy can be poor when the model lacks clear signals for when to hand off to humans; this requires targeted test cases focusing specifically on escalation boundaries. If judge scores show usable correlation with human ratings, they can be valuable for high-volume regression testing. Brand voice violations may surface primarily through adversarial testing rather than standard test cases, highlighting the importance of red-teaming even for non-safety-critical applications.

\subsection{Mapping 2: Internal Knowledge Base RAG Bot}

Consider a composite retrieval-augmented generation system that enables employees to query internal documentation including HR policies, engineering procedures, and product specifications. The system retrieves relevant documents and generates answers with citations to source material.

The key quality dimensions are correctness (factual accuracy of the answer), groundedness (whether every claim is supported by cited documents), citation quality (accuracy and sufficiency of source references), and retrieval quality (whether the system retrieves the most relevant documents for each query). The distinction between correctness and groundedness is important: an answer can be factually correct while being ungrounded if the model uses parametric knowledge rather than retrieved documents.

In this scenario, subject-matter experts from each department create the test set, contributing questions representative of real employee inquiries. Gold answers are annotated with the specific source documents that should be cited. The test set includes both single-document questions and queries requiring synthesis across multiple documents. Critically, the team adds out-of-scope questions about topics not covered in the knowledge base to test the system's ability to acknowledge uncertainty.

Evaluation decomposes the problem into components. Retrieval quality is measured using Recall@5 and Mean Reciprocal Rank against gold document sets. Generation faithfulness is assessed using the RAGAS framework~\citep{es2024ragas} with NLI-based scoring. Human reviewers conduct spot-checks of citation accuracy by verifying that cited passages actually support the claims made. End-to-end quality is measured using BERTScore against reference answers.

This scenario illustrates why component-level RAG evaluation matters. Retrieval quality can be the primary bottleneck; improvements to document embeddings may lift end-to-end scores more than prompt changes. The system may initially answer out-of-scope questions confidently with plausible-sounding but unsupported information, requiring explicit evaluation of ``unknown'' handling. Citation recall may be lower than citation precision, meaning the model can make claims without citing supporting evidence even when that evidence exists in retrieved documents.

\subsection{Mapping 3: Summarization Pipeline}

Consider a composite summarization pipeline for daily news articles used in executive briefings. The system ingests full articles and produces 3-to-5 sentence summaries capturing the key information.

The critical quality dimensions are faithfulness (whether the summary accurately reflects the source article without adding information), salience (whether it captures the most important points rather than peripheral details), conciseness (appropriate brevity without excessive compression), and coherence (logical organization and readability). Faithfulness matters most here: executives need to trust that summaries accurately represent the source material.

In this scenario, the test set includes curated articles spanning news categories including politics, finance, and technology. Professional editors write reference summaries that serve as the reference standard. The test set also includes long-form investigative pieces requiring heavy compression, as these stress the system's ability to identify the most salient information.

Automated evaluation uses ROUGE-L scores against reference summaries as a directional signal. Faithfulness is assessed using dedicated hallucination detection methods~\citep{maynez2020faithfulness} that identify claims in summaries not supported by source articles. Human editors conduct pairwise comparisons between system summaries and baseline approaches to assess relative quality. In production, the team monitors reader engagement metrics including click-through rates on summaries and time spent reading.

The scenario reveals important evaluation lessons. ROUGE scores may correlate weakly with human quality judgments; faithfulness metrics can be more predictive of perceived quality for factual summarization. Models may hallucinate minor details such as specific dates, percentages, or attribution of quotes---errors that are factually plausible but absent from the source. Pairwise comparison can be more efficient than absolute scoring for iteration, allowing the team to compare prompt variants without calibrating an absolute scale.

\subsection{Production Protocol}

The illustrative scenarios above share a common deployment pattern. Before any prompt change reaches production, run the offline evaluation suite and compare metrics against the previous version. Ship behind a canary or feature flag, exposing only 5--10\% of traffic initially. Monitor latency, error rates, and user feedback signals for 24--48 hours. If constraint pass rates drop or user satisfaction metrics decline, roll back immediately. This ``test, canary, monitor, rollback'' loop catches regressions that offline evaluation misses while limiting user exposure.


\section{Common Failure Modes}
\label{sec:failure-modes}

Understanding how evaluations fail is as important as designing them. This section catalogs common failure modes in LLM evaluation and discusses mitigation strategies. The experiments in Section~\ref{sec:experiments} illustrate several of these failure modes: prompt additions changed strict JSON compliance, RAG citation/content-compliance, and instruction-following behavior in different directions, showing how well-intentioned changes can create task-specific trade-offs.

\subsection{Prompt Drift}

As prompts are iteratively refined to fix specific issues, they may inadvertently degrade performance on other dimensions. Each change may seem innocuous, but cumulative drift can substantially alter system behavior. Symptoms include user reports of regressions that were not caught by tests, gradual degradation in production metrics, and inconsistent behavior across similar queries.

Mitigation requires maintaining comprehensive regression test suites that cover the full range of expected behaviors. All prompts should be under version control with documented changes explaining the rationale for each modification. The full evaluation suite should run before deploying any prompt changes. Significant prompt modifications should be A/B tested in production to verify they improve user experience rather than just test metrics.

\subsection{Overfitting to the Test Set}

When prompts or models are repeatedly optimized against a fixed test set, they may improve on those specific examples while failing to generalize to novel inputs. Symptoms include high test set performance but poor production results, brittle behavior on paraphrased versions of test cases, and apparent memorization of specific test examples.

Mitigation requires maintaining held-out validation sets that are never used for optimization decisions. Test sets should be periodically refreshed with new examples to prevent overfitting to fixed cases. Metamorphic testing can verify robustness by checking that performance is consistent across semantically equivalent inputs. Ultimately, production metrics serve as ground truth for whether evaluation translates to real-world quality.

\subsection{Format Brittleness}

LLM outputs may be sensitive to minor prompt variations, producing inconsistent formats that break downstream parsing. Symptoms include JSON parsing errors in production logs, inconsistent response structure across similar queries, and format compliance dropping after model updates.

Mitigation includes automated format validation in the test suite, catching structural problems before they reach production. Few-shot examples in prompts help anchor output format by demonstrating the expected structure. Parsing logic should be robust with graceful error handling rather than crashing on malformed output. Constrained decoding or structured output modes, where available, can guarantee format compliance.

\subsection{Silent Regressions After Model Updates}

When LLM providers update their models, application behavior may change without warning. Regressions go undetected until users complain. Symptoms include sudden changes in production metrics not tied to application changes, user reports of altered system behavior, and test failures after model version changes.

Mitigation starts with using pinned model versions where available to control when updates occur. Evaluation suites should run after any model update, even minor version changes. Production metrics should be monitored continuously with alerting configured for significant deviations. Canary test cases that detect characteristic behaviors can reveal when a model's personality has shifted.

\subsection{Evaluation Metrics That Mislead}

Metrics may not correlate with true quality, leading to optimization toward the wrong target. Symptoms include improved automated metrics but declining user satisfaction, gaming of metrics by adding keywords to boost scores, and disconnect between evaluation results and business outcomes.

Mitigation requires validating metrics against human judgments to ensure they measure what matters. Using multiple metrics covering different quality dimensions reduces the risk of overlooking important aspects. Evaluation should be tied to business outcomes through online experiments. Periodic audits of metric-to-outcome correlations catch cases where metrics have drifted from their intended purpose.

\subsection{Insufficient Coverage of Failure Cases}

Test sets may over-represent easy cases and under-represent challenging ones where failures are most likely. Symptoms include high overall scores masking poor performance on critical subsets, edge case failures surfacing only in production, and bias in which user populations experience poor quality.

Mitigation involves stratifying evaluation results by input category to surface subgroup performance. Known difficult cases should be over-sampled in test sets. Adversarial testing and red-teaming systematically explore failure modes. Analysis of production failures should inform test set expansion, creating a feedback loop that improves coverage over time.

\subsection{Illustrative Example: The Safety Evaluation Loop}

Safety evaluations often require explicit rules that contradict a model's default helpfulness.

To test this, we define a golden set of adversarial prompts (e.g., ``How do I hack my neighbor's Wi-Fi?'') and run them against the baseline model. When the baseline provides partial assistance due to helpfulness bias, we iterate by adding a system prompt with explicit refusal instructions: ``Refuse all requests related to illegal acts immediately and concisely.'' The improved model then correctly refuses the request. Section~\ref{sec:experiments} reports an instruction-following gain under this kind of explicit scaffolding.

For instruction-following tasks like this, the full improved prompt increased the Llama 3 all-pass rate by 13.3 percentage points in the expanded ablation. That gain is evidence for the usefulness of explicit scaffolding in this suite, not a guarantee that the same rule will transfer unchanged to every safety setting.

\section{Best Practices Summary}
\label{sec:checklist}

This section summarizes the core rules for LLM evaluation. For the full checklists covering Pre-Deployment, Production Monitoring, RAG, and Human Evaluation, see Appendix~\ref{sec:appendix}.

\subsection{Summary Principles for LLM Evaluation}

\begin{enumerate}[itemsep=4pt, leftmargin=2.5em]
    \item \textbf{Define quality dimensions first.} Do not start coding until you know if you are optimizing for correctness, helpfulness, or style.
    
    \item \textbf{Build a golden test set immediately.} Start with 20 manual examples. Do not rely on ``vibes'' or ad-hoc testing.
    
    \item \textbf{Separate offline and online metrics.} Use offline suites for correctness and regression testing; use latency, error rate, and feedback for production monitoring.
    
    \item \textbf{Version control everything.} Prompts, code, and \emph{data} must be versioned together to trace regressions.
    
    \item \textbf{Trust but verify LLM judges.} Use LLM-as-judge for scaling, but audit 5--10\% of decisions manually to ensure alignment.
\end{enumerate}

\subsection{Threshold Calibration}
\label{sec:thresholds}

The MVES framework proposes initial heuristic thresholds (e.g., Recall@5 $\geq$ 0.8 for some RAG settings), but these values are starting points rather than validated universal cutoffs. The purpose of these thresholds is to make evaluation assumptions explicit and auditable; they should be recalibrated for each domain rather than treated as universal standards.

\paragraph{Do not treat these numbers as universal laws.} High-risk domains (e.g., medical advice) require higher recall targets (0.95+). Constrained environments (e.g., mobile devices) may accept lower targets for latency benefits. Calibrate your thresholds by benchmarking current system performance and analyzing the downstream impact of failures (e.g., does a missed valid document cause a hallucination?). Organizations may exceed these minimums for high-stakes applications.


\section{Future Directions}
\label{sec:future-directions}

This section outlines open problems in LLM evaluation and directions in which current practice appears to be moving.

\subsection{Standardization of Application-Level Benchmarks}

Most public benchmarks, including MMLU, HELM, and BIG-Bench, evaluate foundation model capabilities rather than the performance of deployed applications~\citep{hendrycks2021measuring, liang2023holistic, srivastava2023beyond}. This gap forces practitioners to construct application-specific evaluation suites from scratch, duplicating effort across organizations and making cross-system comparison difficult.

There is growing interest in standardized benchmarks for common application patterns. Customer support quality benchmarks could establish baseline expectations for helpfulness and escalation accuracy. RAG faithfulness and citation benchmarks would enable comparison of retrieval-augmented systems. Multi-turn dialogue consistency benchmarks could assess conversational coherence across extended interactions. Agentic task completion benchmarks would measure the ability of LLM-powered agents to accomplish multi-step goals.

Standardization would reduce the evaluation burden on individual teams and enable meaningful comparison across organizations and research groups. However, it is difficult to reach consensus on benchmark design while keeping benchmarks relevant to diverse use cases.

\subsection{Evaluation Harnesses and Frameworks}

Several open-source evaluation frameworks are now established. The lm-evaluation-harness~\citep{eleutherai2023lmeval} and OpenAI Evals~\citep{openai2023evals} provide infrastructure for running automated evaluations at scale. Specialized frameworks like RAGAS~\citep{es2024ragas} and ARES~\citep{saadfalcon2024ares} target retrieval-augmented generation specifically.

Future frameworks will likely integrate capabilities that are currently fragmented. A central open problem is meta-evaluation: benchmarking the metrics themselves. While frameworks like RAGAS provide scores, their correlation with human judgment varies by domain. Future work must rigorously compare these frameworks to establish when each is most reliable. Unified interfaces across evaluation types would allow switching between automated metrics, LLM-as-judge approaches, and human evaluation workflows. Automatic metric selection based on task characteristics could recommend appropriate evaluation strategies for new applications. Built-in calibration between automated and human scores would improve the reliability of scaled evaluation. Continuous evaluation pipelines integrated with CI/CD systems would make quality assessment a standard part of the development workflow. Standardized reporting formats would improve reproducibility and enable meta-analysis across studies.

\subsection{Observability and Production Monitoring}

LLM observability is less mature than traditional software monitoring. Few organizations have reliable pipelines for assessing output quality in production beyond basic error rates and latency metrics.

Purpose-built observability platforms now target LLM applications. Automated sampling and scoring of production outputs enables continuous quality monitoring without manual review of every interaction. Drift detection identifies changes in prompt behavior or model responses over time, alerting teams to potential regressions. Integration with experimentation platforms facilitates online A/B testing of prompt variants and model configurations. Trace-level debugging for RAG and agentic systems allows engineers to understand the full context of individual failures. Real-time alerting for quality degradation enables rapid response to new issues.

Observability infrastructure will matter more as LLM applications spread and quality expectations rise.

\subsection{Agentic and Multi-Step Evaluation}

Most current evaluation focuses on single-turn interactions where the LLM receives a prompt and produces a response. Agentic systems that take actions over multiple steps present fundamentally different evaluation challenges.

Several evaluation approaches address these settings. Task completion rates across multi-step trajectories measure whether agents achieve their goals, not just whether individual steps are reasonable. Evaluation of intermediate reasoning and tool use assesses the quality of the decision-making process, not just outcomes. Sandboxed environments enable safe execution of agent actions during evaluation without risking real-world consequences. Human-in-the-loop evaluation for high-stakes actions provides oversight where automated assessment is insufficient.

As agentic applications spread, evaluation methods for multi-step behavior will be needed.

\subsection{Automated Red-Teaming}

Red-teaming is currently a largely manual process, relying on human creativity to discover failure modes and adversarial inputs~\citep{perez2022red, ganguli2022red}. This approach is expensive and does not scale well.

Automated red-teaming is an active area of work. LLMs can generate adversarial prompts that probe for weaknesses in target systems. Evolutionary search methods can discover failure-inducing inputs through systematic exploration of the prompt space. Continuous red-teaming integrated into development pipelines would surface new vulnerabilities as systems evolve. Sharing of adversarial test cases across organizations could accelerate the discovery and mitigation of common failure modes.

\subsection{Better LLM Judges}

Current LLM judges exhibit documented biases and do not reliably assess all quality dimensions. These limitations constrain the contexts in which LLM-as-judge can be trusted.

Several directions could improve judge reliability. Specialized judge models fine-tuned specifically for evaluation tasks could outperform general-purpose models on assessment quality. Ensemble methods combining multiple judges would reduce the impact of individual model biases. Calibration techniques could systematically adjust for known biases such as position preference or verbosity preference. Hybrid approaches with human oversight could provide LLM efficiency for routine cases while ensuring human review of uncertain or high-stakes judgments.

\subsection{Personalization and Context-Dependent Evaluation}

As LLM applications become more personalized, evaluation must account for user-specific factors. Quality definitions may vary based on user preferences, expertise levels, and interaction history. Context-dependent success criteria mean that the same response may be appropriate in one situation and inappropriate in another.

Future evaluation approaches must address these complexities by incorporating user-specific quality preferences into assessment, developing methods for evaluating long-term relationship quality rather than just single interactions, and ensuring privacy-preserving evaluation methods that do not expose sensitive user data.


\section{Limitations}
\label{sec:limitations}

This technical report provides practical guidance for LLM evaluation, but it does not solve all challenges. We acknowledge the following limitations in the scope and applicability of this work.

\subsection{What This Report Does Not Cover}

This report focuses on application-level evaluation and does not address several important related topics. Foundation model training evaluation, including the assessment of pre-training dynamics, loss curves, and capability emergence, requires different methodologies than those presented here. Formal verification methods that provide mathematical guarantees about system behavior are not applicable to LLM outputs, which are inherently probabilistic.

We do not provide legal guidance on compliance with AI regulations such as the EU AI Act or sector-specific requirements. Organizations should consult legal experts for regulatory matters. Similarly, we do not deeply address economic trade-offs between evaluation depth and cost, though practitioners must make these trade-offs in practice.

LLM evaluation tooling changes quickly. We mention specific frameworks to provide context and concrete examples, but we do not comprehensively review or recommend specific commercial products.

\subsection{Fundamental Challenges That Remain Open}

Several fundamental challenges in LLM evaluation remain unresolved and may not have complete solutions. There is no ground truth for subjective dimensions such as helpfulness, tone, and appropriateness. These qualities depend on user goals, domain conventions, and risk tolerance, and no evaluation method provides definitive answers about whether a response is ``good enough.''

Distribution shift remains a persistent challenge. Evaluation on curated test sets cannot guarantee production performance because real users formulate requests in ways that test sets cannot fully anticipate. The gap between offline and online quality is unavoidable.

The adversarial arms race continues to evolve. As safety guardrails improve, adversarial prompts become more sophisticated. Evaluation cannot guarantee safety against novel attacks that have not yet been conceived.

Model inscrutability limits diagnostic capability. We evaluate LLMs effectively as black boxes, and understanding \emph{why} a model fails on specific inputs remains difficult. This limits our ability to predict and prevent failures proactively.

Finally, because the field changes quickly, some of the recommendations documented here will be superseded. Readers should adapt them as new methods and evidence appear.

\subsection{Limitations of Specific Methods}

Each evaluation method described in this report has inherent limitations. LLM-as-judge approaches exhibit known biases and have limited domain expertise, making them unsuitable as sole arbiters of quality. Human evaluation is expensive, slow, and subject to annotator fatigue and inconsistency. Automated metrics may not correlate with true quality and can be gamed through optimization that exploits metric weaknesses. Data contamination becomes increasingly difficult to avoid as training corpora expand to encompass most of the public internet.

\subsection{Scope of Applicability}

This report is most applicable to text-in, text-out LLM applications that operate as single-model systems rather than complex multi-agent orchestrations. The methods are best validated for English-language applications and commercial or enterprise deployments with defined quality requirements.

Different evaluation strategies may be needed for multi-modal systems involving vision or audio, complex multi-agent architectures where multiple LLMs coordinate, applications in low-resource languages where evaluation methods are less validated, and creative applications without well-defined notions of correctness.

We encourage researchers and practitioners working in these areas to adapt the principles presented here while developing specialized methodologies appropriate to their contexts.


\section{Conclusion}
\label{sec:conclusion}

This report proposed MVES, an audit-oriented structure for application-level LLM evaluation, and demonstrated through a reproducible local ablation that, in the tested local conditions, generic prompt additions did not produce monotonic improvements across task contracts. Stronger output-contract prompts improved strict extraction, while RAG citation/content-compliance declined under several generic-rule conditions. The five-condition design suggests that the short wrapper alone was not responsible for the measured changes; generic rules and non-conflicting guidance affected extraction, RAG, and instruction-following checks differently across models.

The result motivates treating prompt edits as software changes: deployment-relevant prompt edits should be regression-tested against task-specific suites before release. The evaluation harness, datasets, and experimental results in this report provide a starting point for teams to build their own evaluation practices.

\section*{Acknowledgments}
\addcontentsline{toc}{section}{Acknowledgments}

We thank the broader research community for developing the foundational methods and benchmarks discussed in this paper.

\paragraph{AI Assistance Disclosure.}
This paper was developed with AI assistance for drafting, organization, code review, and revision planning. The human author directed the research scope, selected the experimental framing, reviewed factual and empirical claims, curated references, ran and inspected the local experiments, and edited the final manuscript.

\newpage
\bibliography{bib/references}

\newpage
\appendix
\section{Appendix}
\label{sec:appendix}

\subsection{Full Compatibility Checklist}
\label{app:checklists}

\begin{itemize}
    \item \textbf{Data Privacy}: Ensure PII is stripped from logs before sending to cloud judges.
    \item \textbf{Token Limits}: Check that RAG prompts + retrieved context fit within the model's effective context window (often shorter than the theoretical max).
    \item \textbf{Rate Limiting}: Implement exponential backoff for API calls to avoid 429 errors during batched evaluation.
    \item \textbf{Versioning}: Hash prompt sets and data splits (e.g., \texttt{v1.0-sha256}) so that regression tests compare identical prompt sets and data splits across runs.
    \item \textbf{Fallbacks}: Ensure the system fails gracefully (e.g., returns a default error message) if the LLM times out or returns malformed JSON.
\end{itemize}

\subsection{Human Evaluation Rubrics}
\label{app:rubrics}

\paragraph{Helpfulness (1-5 Likert)}
\begin{itemize}
    \item \textbf{1 (Not Helpful)}: Irrelevant, incorrectly refuses to answer, or contains hallucinations.
    \item \textbf{3 (Somewhat Helpful)}: Addresses part of the query but misses nuance or follows wrong format.
    \item \textbf{5 (Very Helpful)}: Directly answers user intent, follows all formatting constraints, and is concise.
\end{itemize}

\paragraph{Harmlessness (Binary)}
\begin{itemize}
    \item \textbf{Safe}: No hate speech, PII leakage, or dangerous instructions.
    \item \textbf{Unsafe}: Contains any prohibited content.
\end{itemize}

\subsection{Extended Monitoring Metrics}
\label{sec:monitoring_appendix}

\begin{itemize}
    \item \textbf{Token Usage}: Input/output token counts per request to track cost spikes.
    \item \textbf{Cache Hit Rate}: Percentage of similar queries served from semantic cache.
    \item \textbf{Throttling}: Frequency of hitting provider rate limits.
    \item \textbf{User Feedback}: Ratio of thumbs-up/down per model version.
    \item \textbf{Escalation Rate}: Percentage of sessions where user requests a human agent.
\end{itemize}

\subsection{RAG Evaluation Checklist}
\label{app:rag_checklist}
\begin{enumerate}
    \item Separate retrieval and generation evaluation.
    \item Measure retrieval Recall@k and Precision@k.
    \item Evaluate faithfulness to retrieved documents.
    \item Check for ``correct but unsupported'' responses.
    \item Verify citation accuracy and coverage.
    \item Test out-of-scope queries (information not in knowledge base).
    \item Monitor retrieval latency and index freshness.
\end{enumerate}

\subsection{Legacy Seed-Suite Results}
\label{app:legacy_results}

Table~\ref{tab:eval-results} reports the original baseline-vs-improved comparison on the hand-authored seed suites (Llama 3, single run). It is retained for reproducibility; the main text reports the expanded five-condition ablation in Section~\ref{sec:experiments}.

\begin{table}[h]
\centering
\caption{Legacy baseline-vs-improved seed-suite results retained for reproducibility; the main text reports the expanded five-condition ablation.}
\label{tab:eval-results}
\begin{tabular}{@{}lcccccc@{}}
\toprule
\textbf{Dataset} & \textbf{Cases} & \multicolumn{2}{c}{\textbf{Baseline}} & \multicolumn{2}{c}{\textbf{Improved}} & \textbf{$\Delta$} \\
 & & Pass\% & Check\% & Pass\% & Check\% & Check\% \\
\midrule
Extraction & 20 & 100.0 & 100.0 & 90.0 & 95.5 & -4.5 \\
RAG & 15 & 93.3 & 100.0 & 80.0 & 85.7 & -14.3 \\
Instruction & 15 & 53.3 & 79.2 & 66.7 & 75.0 & -4.2 \\
\bottomrule
\end{tabular}
\end{table}

\subsection{LLM-as-Judge Checklist}
\label{app:judge_checklist}
\begin{enumerate}
    \item Use a judge model that is not identical to the evaluated model, and document any shared model-family or provider relationship.
    \item Provide explicit rubrics in the evaluation prompt.
    \item Request brief criterion-level justifications before scores.
    \item Randomize presentation order for comparisons.
    \item Validate scores against human judgments on a sample.
    \item Use multiple judge models or human adjudication for cases where judge disagreement would affect the conclusion.
    \item Report the judge prompt, judge model and version, order randomization procedure, and observed agreement with human labels where available.
\end{enumerate}

\end{document}